%% file: main.tex
\documentclass[11pt]{article}
\PassOptionsToPackage{dvipsnames,table}{xcolor}
\usepackage{smacros_arxiv}

\usepackage[numbers,square]{natbib}
\usepackage{algorithmic}
\usepackage{mathbbol}
\usepackage{placeins}
\usepackage{float}
\usepackage{multirow}
\usepackage{xurl}

\makeatletter
\@ifundefined{theHtheorem}{}{}
\@ifundefined{theHlemma}{}{}
\@ifundefined{theHremark}{}{}
\@ifundefined{theHproposition}{}{}
\makeatother
\definecolor{goodgreen}{RGB}{220,245,220}
\definecolor{badred}{RGB}{250,220,220}
\definecolor{bestblue}{RGB}{220,230,250}
\title{Risk-Sensitive Non-Parametric and Kernel Bandits}

\author{%
  Devdan Dey 
 \footnote{ Department of Computer Science and Engineering,
  Indian Institute of Technology Bombay.
  Email: \url{23d0365@iitb.ac.in}} \quad \quad
  Sujoy Bhore 
  \footnote{Department of Computer Science and Engineering,
  Indian Institute of Technology Bombay.
  Email: \url{sujoy@cse.iitb.ac.in}} \quad \quad
  Avishek Ghosh
 \footnote{ Department of Computer Science and Engineering,
  Indian Institute of Technology Bombay.
  Email:
  \url{avishek_ghosh@iitb.ac.in}}
}

\date{}

\begin{document}
\maketitle

\begin{abstract}
We study the non-parametric bandit problem where actions are evaluated using
\emph{Cumulative Prospect Theory (CPT)} rather than expected reward. We consider
the above problem over continuous action spaces with arm-dependent noise. We
propose and analyze \emph{Non-Parametric CPT Bandits} (\texttt{NPCB}), which
chooses the actions in an optimistic fashion with CPT weights. Notably,
\texttt{NPCB} decouples the mean estimation task from CPT weight-based decision
making, which makes it \emph{modular}. We show that the appropriately defined
(CPT) regret of \texttt{NPCB} is
$\widetilde{\mathcal O}\!\left(
\sqrt{T\,\dim_E(\mathcal F,T^{-2})\,
\log N(\mathcal F,T^{-2},\|\cdot\|_\infty)}
\right)$, where $\dim_E$ is the \textit{Eluder}-dimension of the
non-parametric function class $\mathcal F$, $T$ is the learning horizon and
$N(\mathcal F,T^{-2},\|\cdot\|_\infty)$ is the covering number of $\mathcal F$
in the sup norm with precision $T^{-2}$. This matches the
usual regret of classical non-parametric bandits
\cite{DBLP:conf/nips/RussoR13}, which proves that
adding risk-sensitiveness to the formulation is order-wise \emph{free}.

{We further extend \texttt{NPCB} to unknown noise CDFs. The regret
bound retains the known-CDF mean-learning term and adds a CDF-learning cost
governed by a distributional Eluder dimension.}
As an instantiation, we study \emph{Kernel-CPT} bandits.
{Using information gain, we obtain regret
bounds for a large variety of kernels. This bound matches the corresponding
risk-neutral kernel upper rate up to logarithmic factors.}
\end{abstract}

\input{paper}

\section*{AI Use Statement}

In this work, we used generative AI tools to write the Python
code for the experiments and plots in
Section~\ref{sec:experiments} and
Appendix~\ref{app:experiments}.
The experimental design, including which experiments to run
and which questions to investigate, was conceived entirely
by the authors, and the generated code was manually verified
for correctness by one of the authors.
Additionally, we used generative AI tools for editing and
grammar-checking the manuscript text. 
We have reviewed all AI-assisted work and take responsibility
for the final content of this work, including text, claims,
and artifacts.

\bibliographystyle{alphaabbrv}
\bibliography{ref}

\clearpage
\appendix
\input{appendix}
\end{document}

%% file: paper.tex
\section{Introduction}
\label{sec:intro}
The Multi-Armed Bandit framework is a fundamental paradigm for sequential decision-making under uncertainty. In the vast majority of this literature \cite{lattimore2020bandit,bubeck2012regret}, the objective is to minimize \textit{appropriately} defined regret, which depends simply on the mean of the reward distribution associated with the arm pulled. However, in real-world applications—ranging from clinical trials to financial portfolio optimization—the mean does not satisfactorily capture the merits of certain actions, especially when humans are involved in the decision. See \cite{tan2022surveyriskawaremultiarmedbandits} for a recent survey.

In such applications, risk plays an important role, and a risk-aware performance measure is often preferred over the average reward. Consequently, significant research has focused on diverse risk measures, including mean-variance (MV) \cite{e5a1bb8f-41b7-35c6-95cd-8b366d3e99bc}, Value-at-Risk (VaR) \cite{https://doi.org/10.1111/1467-9965.00068}, Conditional Value-at-Risk (CVaR) \cite{https://doi.org/10.1111/1467-9965.00068, Rockafellar2000OptimizationOC}, spectral risk measures \cite{ACERBI20021505}, and utility-based shortfall risk \cite{https://doi.org/10.1111/1467-9965.00068, Hu2018UtilitybasedSR}. While these measures satisfy axioms of coherence or convexity, they often fail to model the nuances of human behavior, such as loss aversion and asymmetric preferences, which are best captured by the celebrated (and Nobel prize-winning) risk measure Cumulative Prospect Theory (CPT) \cite{tversky1992advances}. The failure of standard risk measures is structural. MV penalizes all variance
equally, treating rare massive gains and catastrophic losses identically. CVaR only examines the lower tail of the reward distribution, so
rare disasters are averaged with ordinary outcomes when their
probability is below the chosen tail level, while attractive
lottery gains are not its focus. CPT avoids both failures through nonlinear probability
weighting --- a $1\%$ chance of a fatal crash carries disproportionate
psychological weight far exceeding $0.01$. This amplification of rare events
means CPT simultaneously penalizes hidden disasters and values rare
large gains, capturing the two asymmetric behaviors that characterize human
decision-making under uncertainty.

To further motivate the necessity of CPT risk, consider the challenge of automated policy design motivated by the \emph{Asian Disease Problem} \cite{Kahneman2011-KAHTFA-2}. Consider an agent tasked with optimizing a Public Health Score $r$ (e.g., the number of lives saved). The agent must choose between two policies: a ``Safe'' protocol guaranteeing that 200 lives are saved ($r=200$), and a ``Risky'' protocol with a 35\% chance of saving 600 lives ($r=600$) and a 65\% chance of saving none ($r=0$). A standard risk-neutral agent calculates the averages as $ \mu_{\text{Risky}} = 210 > \mu_{\text{Safe}} = 200$, therefore identifying the risky protocol as \textit {optimal}. However, empirical evidence shows that the vast majority of human decision-makers reject this gamble \cite{tversky1992advances} and choose the safe protocol. This reveals a critical insight: the standard method of maximizing the mean reward is misaligned with human ethical and behavioral preferences.

CPT addresses this misalignment by evaluating actions not through a simple average, but through a distorted value $\mathcal{V}_a(f) := \int_{0}^{\infty} \mathbb w(1 - F^{f}_{a}(z) ) dz -\int_{0}^{\infty} \mathbb w\left(F^{f}_a(-z) \right) dz$, where $\mathbb w(\cdot)$ is a non-linear weight (distortion) function applied to the cumulative distribution $F_X(.)$. Note that if $\mathbb w(\cdot)$ is an identity function, we get the standard average, i.e., $\mathcal{V}(X) = \mathbb{E}(X)$. In the above-mentioned example, $\mathcal{V}_{\text{Safe}}=200$ while
$\mathcal{V}_{\text{Risky}}=600\,\mathbb w(0.35)$. Hence the safe action is
the subjective optimum precisely when $\mathbb w(0.35)<1/3$. For instance, using the
Tversky--Kahneman weighting form with $\gamma=0.5$ gives
$\mathbb w(0.35)\approx 0.303$, and therefore
$\mathcal{V}_{\text{Risky}}\approx 182<200=\mathcal{V}_{\text{Safe}}$. In general, with distorted weights, the sequence of actions chosen may differ from that under the risk-neutral measure.
To demonstrate this distinction empirically, we construct instances in which
risk-neutral and risk-sensitive objectives lead to systematically different
decisions, with the separation persisting across repeated trials
(see, Figure~\ref{fig:main_risk_kernel} (a)-(b)).

Significant research has focused on diverse risk measures in the multi-armed bandit setting. For instance, \cite{DBLP:journals/corr/abs-1301-1936, DBLP:journals/corr/VakiliZ16,hu2024multiarmedbanditsproblemmeanvariance} study the Mean-Variance setup extensively for $K$-armed bandits.
\cite{7447162} analyzes Value-at-Risk (VaR) in the context of the total reward sequence.
\cite{ DBLP:journals/corr/GalichetST14, Tamkin2019DistributionallyAwareEF,Tan2022CVaRRegretBF,DBLP:journals/corr/abs-1901-00997} utilize the Conditional Value-at-Risk (CVaR) measure in $K$- armed Bandit setting. \cite{DBLP:journals/corr/abs-1906-00569} proposes a distribution-oblivious algorithm for optimizing a linear combination of mean and CVaR, specifically for unbounded reward distributions in the best-arm identification setting. For Kernel bandits, \cite{DBLP:conf/icml/Takemori22} study the standard Upper Confidence Bound (UCB) type algorithms for CVaR and Mean-Variance risk measures and prove near-optimal results. To provide a systematic analysis of such metrics, \cite{pmlr-v75-cassel18a} introduced a unified framework for multi-armed bandits under general risk satisfying a Lipschitz continuity condition with respect to probability distributions (e.g., in the Wasserstein distance). However, the CPT functional is generally \emph{not} Lipschitz continuous \cite{tan2022surveyriskawaremultiarmedbandits}, often causing the value to fluctuate wildly with small distribution changes. This lack of stability prevents the direct application of existing unified frameworks to CPT.

To the best of our knowledge, bandit optimization with CPT-based risk measures has been limited. Prior works have used CPT-based risk-sensitive measures for simple $K$-armed and linear bandits \cite{DBLP:journals/corr/GopalanAFM16}, and achieve near-optimal regret bounds. Recently, \cite{pmlr-v258-tatli25a} proposes a unified framework for
distortion risk measures in the $K$-armed bandit setting, covering both
monotone and non-monotone distortions and showing that, for certain
non-monotone objectives, an optimal policy may require a mixture of arms.
However, their analysis is primarily focused on finite $K$-armed bandits,
whereas our interest is in continuous action spaces and general
non-parametric reward functions. Moreover, CPT introduces a distinct
nonlinear probability distortion that is not covered by their mixture-based
analysis. This leaves a theoretical gap in CPT optimization for
non-parametric function classes. This prompts the central question of our
work:

\begin{center}
    
    \emph{\textit{Is it possible to learn (general) non-parametric
    (including Kernel) bandits with CPT risk measure?}}
    
\end{center}

In this paper, we answer this question affirmatively. We analyze a learning algorithm, namely Non-Parametric CPT-Bandits (\texttt{NPCB}), that works for general functional bandits with the CPT risk measure. Notably, \texttt{NPCB} is agnostic to  (non-parametric) function class and uses the \textit{Eluder Dimension} \cite{DBLP:conf/nips/RussoR13} to characterize the complexity of the learning task. As a primary instantiation of our framework, we study Kernel bandits and obtain guarantees for \textit{Kernel CPT} (\texttt{KCB}) algorithm. To the best of our knowledge, this is the first work to study Kernel Bandits (RKHS) under a CPT-based risk measure.

\subsection{Summary of contributions}
\textbf{Non-parametric CPT bandits:} We propose a learning algorithm with CPT-based risk for general, non-parametric bandits, namely \texttt{NPCB}. Our formulation captures continuous action spaces with arm-dependent noise. For this general setup, we derive the regret bound (refer to Theorem \ref{thm:general_regret}) which matches the order-wise performance of risk-neutral non-parametric bandits \cite{DBLP:conf/nips/RussoR13}. This demonstrates that non-linear human-centric risk measures can be incorporated into infinite-dimensional function learning without incurring a regret penalty.

\textbf{Kernel CPT bandits:}
For an RKHS mean-reward class and known arm-dependent noise CDFs,
we give kernel algorithms for CPT regret.
Our phasewise procedure achieves
$\widetilde O(\sqrt{T\gamma_T(\lambda)})$ regret,
matching the standard risk-neutral dependence on information-gain.

\textbf{Modular Algorithm:} Our algorithms decouple the \textit{Learning Module} (mean estimation from undistorted rewards) from the \textit{Decision Module} (CPT valuation). This makes our framework function-class-agnostic, allowing it to be seamlessly applied to any function class (e.g., Linear, RKHS, or Neural Nets), provided a valid confidence set can be constructed.

\textbf{CPT stability for non-parametric learning:}
Building on the translation stability
of~\cite{DBLP:journals/corr/GopalanAFM16},
we connect CPT uncertainty to confidence widths for general
mean-function classes.
Under additive noise, the CPT value is uniformly $2$-Lipschitz
in the mean, even without H\"{o}lder continuity of the weighting function \footnote{The finite-arm analysis
of~\cite{DBLP:journals/corr/GopalanAFM16}
requires H\"older continuity of $\mathbb w$.
Their known-CDF linear analysis avoids this requirement.
Our framework extends this property to general mean-function classes
while retaining the corresponding risk-neutral complexity dependence.}.
This connects CPT regret to the Eluder dimension and covering numbers,
and supports our kernel instantiations.

\textbf{Removing Known CDF assumption:} For the case of linear bandits,
\cite[Section~V, Remark~10]{kolla2023banditalgorithmsemulatehuman} require the knowledge of the cumulative distribution function of the noise, remarking that online estimation of the CDF is an open problem left for future work. {We learn the noise CDF online using a joint likelihood
confidence set. The regret keeps the mean-learning term and adds a
cost governed by joint distributional complexities. We bound these
complexities for an explicit linear family and its RKHS extension.}

\section{Problem Formulation}
\label{sec:formulation}

We consider the sequential decision-making problem over a finite time horizon $T$, characterized by a compact and continuous action space $\mathcal{A}\subseteq \mathbb{R}^d$, which may be infinite. The environment is governed by an unknown true mean-reward function $f^*: \mathcal{A}\rightarrow \mathbb{R}$, which belongs to a known class of non-parametric functions of real-valued functions $\mathcal{F}$.  The value $f^*(a)$ represents the expected reward associated with action $a \in \mathcal{A}$. At each step $t$, the agent selects an action $a_t \in \mathcal{A}$ based on the history of previous observations $\mathcal{D}_{t-1} = \{(a_\tau, r_\tau)\}_{\tau=1}^{t-1}$. The environment then reveals a realized physical reward
$r_t = f^*(a_t) + \epsilon_t$.

Unlike the standard bandit formulation, we allow \emph{arm-dependent noise}.
Conditional on the history and the selected arm $a_t=a$, the noise
$\epsilon_t$ has a fixed distribution $F_{\epsilon_a}$ with mean zero.
For a candidate $f\in\mathcal F$, the corresponding reward CDF is
$F_a^f(z)=F_{\epsilon_a}(z-f(a)).$
We assume that these candidate reward distributions are supported in
$[-M,M]$.

While the agent observes undistorted rewards $r_t$ drawn from the distribution defined by $F_{\epsilon_{a}}$, its decision is governed by a \emph{subjective} distortion of these probabilities. This distortion is parameterized by weight function $\mathbb w: [0, 1] \to [0, 1]$. We define the weight-distorted reward, denoted by $\mathcal{V}_a(f)$, by applying the distortion $\mathbb w$ to the cumulative probabilities of the reward distribution as
\[\mathcal{V}_a(f) := \int_{0}^{\infty} \mathbb w\left(1 - F^{f}_{a}(z) \right) dz -\int_{0}^{\infty} \mathbb w\left(F^{f}_a(-z) \right) dz\]
The weight function $\mathbb w:[0,1]\to[0,1]$ is the behavioral core of CPT. Beyond
the minimal conditions $\mathbb w(0)=0$, $\mathbb w(1)=1$, and monotone increasing \cite{tversky1992advances,DBLP:journals/corr/GopalanAFM16}, we impose
\emph{no} additional assumptions.
Together with the bounded-reward assumption above, $\mathbb w(0)=0$
ensures that $\mathcal V_a(f)$ is finite.
Several families of $\mathbb w$ satisfying these
properties are widely used in behavioral economics, including the
Tversky--Kahneman function ~\cite{tversky1992advances} $\mathbb w(p) = p^\gamma/(p^\gamma
+ (1-p)^\gamma)^{1/\gamma}$, the Prelec function \cite{prelec1998probability} $\mathbb w(p) = \exp(-(\ln 1/p)^\alpha)$,
and the Gonzalez--Wu function \cite{gonzalez1999shape}$ \mathbb w(p) = \delta p^\gamma/(\delta p^\gamma +
(1-p)^\gamma)$. Our framework accommodates all of these. A detailed discussion
of the behavioral interpretation of $\mathbb w$, its effect on $\mathcal{V}$ is provided in
Appendix~\ref{app:weight_function}.
\begin{remark}
For simplicity, we use reference point $0$, identity utility, and a common
probability weight for gains and losses. The theory extends to a fixed
reference point $x_0$, separate weights $\mathbb w^\pm$, and nondecreasing
$L^\pm$-Lipschitz gain/loss utilities. These extensions change only constant factors, as detailed in Appendix~\ref{app:proof_lipschitz}.
\end{remark}

\paragraph{CPT Regret.}
The agent's objective is to maximize the cumulative subjective, weight-distorted value of the rewards gathered over the horizon. Formally, under the true mean-reward function $f^*$ and distortion mapping $\mathbb w(\cdot)$, we denote the subjectively optimal arm as, $a^* \in \arg\max_{a \in \mathcal{A}} \mathcal{V}_a(f^{*})$. The goal of learning is to maintain a policy that competes with this optimum through sequential interaction with the environment. The performance of the learning policy is evaluated by its \emph{cumulative weight-distorted regret} (CPT regret), given by
\begin{align*}
    R(T) = \sum_{t=1}^T \left( \mathcal{V}_{a^*}(f^{*}) - \mathcal{V}_{a_t}(f^{*}) \right).
\end{align*}

\begin{remark}Designing such an algorithm is challenging because the agent observes samples from the undistorted distribution but must optimize the distorted value $\mathcal{V}_a(f^{*})$. The non-linear transformation $\mathbb w(\cdot)$ prevents the estimation of $\mathcal{V}_a$ through the sample mean of $r_t$. The distortion can inflate the impact of tail events in $F_{\epsilon_{a}}$, potentially changing the preference order between arms.
\end{remark}

\section{Non-parametric CPT bandits (\texttt{NPCB})}
\label{sec:algorithm}

We now present our learning algorithm. Our approach follows the general template of optimistic exploration for general function classes, as discussed by \cite{DBLP:conf/nips/RussoR13}, but fundamentally deviates from it in the arm-selection step to accommodate the agent's behavioral preferences.

\paragraph{Design Philosophy.} At the start of each round $t$, the algorithm constructs a \emph{least squares estimate}, $\hat{f}^{LS}_t \in \operatorname*{arg\, min}_{f \in \mathcal{F}} \sum_{\tau=1}^{t-1} (f(a_\tau) - r_\tau)^2$, which minimizes the cumulative squared prediction error on the history of observations $\mathcal{D}_{t-1}$.
 To quantify the uncertainty around this estimate, we employ the \emph{empirical 2-norm},$\|\cdot\|_{2, \mathcal{D}_t}$. For any function $g$, this is defined as $\|g\|_{2, \mathcal{D}_t}^2 = \sum_{\tau=1}^{t-1} g^2(a_\tau)$. Using this metric, we construct a \emph{confidence set} $\mathcal{C}_t$ centered at the estimate: $\mathcal{C}_t := \left\{ f \in \mathcal{F} : \|f - \hat{f}^{LS}_t\|_{2, \mathcal{D}_{t-1}} \le \sqrt{\beta_t} \right\}.$ Here, $\beta_t$ is an appropriately chosen confidence parameter that ensures the true function $f^*$ lies within $\mathcal{C}_t$ with high probability. The algorithm operates by maintaining this statistically plausible set of environments and selecting actions based on the principle of \emph{Optimism in the Face of Uncertainty}. However, unlike standard algorithms that select the arm with the highest optimistic mean, our algorithm selects the arm  $a_t$ that maximizes the optimistic weight-distorted reward feasible within the confidence set $\mathcal C_t$: $a_t = \operatorname*{arg\, max}_{a \in \mathcal{A}} \sup_{f \in \mathcal{C}_t} \mathcal{V}_a(f).$
The pseudocode for the proposed method is provided in Algorithm \ref{alg:distorted-eluder}.

\paragraph{Modular Design and Function-Class Agnostic Extension.} As mentioned earlier, \texttt{NPCB} is modular, which strictly separates the CPT-based behavioral preferences from the statistical learning of the environment  (e.g., \cite{DBLP:journals/corr/GopalanAFM16}). In Algorithm~\ref{alg:distorted-eluder}, lines 3-4 are for confidence set construction and environment learning, whereas lines 5-6 are used for {CPT valuation and choice of action}.

\paragraph{Computational Complexity.}
We note that for general function classes, solving $\max_{a \in \mathcal{A}}
\sup_{f \in C_t} \mathcal V_a(f)$ is computationally hard in general~\cite{DBLP:conf/nips/RussoR13}.
{For the kernel algorithms, posterior updates use the usual kernel
matrices. We focus on statistical regret and assume access to the
displayed arm maximizers; for the phasewise continuous-domain result,
we use the finite discretization described in Section~\ref{sec:kernel}.}

\begin{algorithm}[t]
\caption{Non-Parametric CPT Bandits (\texttt{NPCB})}
\label{alg:distorted-eluder}
\begin{algorithmic}[1]
\STATE \textbf{Input:} $\mathcal F$, known noise CDFs, and $\{\beta_t\}$.
\FOR{$t=1$ to $T$}
\STATE Set $\hat f_t^{LS}\in\arg\min_{f\in\mathcal F}
\sum_{s<t}(r_s-f(a_s))^2$.
\STATE Set $\mathcal C_t=\{f\in\mathcal F:
\sum_{s<t}(f(a_s)-\hat f_t^{LS}(a_s))^2\le\beta_t\}$.
\STATE Choose $a_t\in\arg\max_{a\in\mathcal A}
\sup_{f\in\mathcal C_t}\mathcal V_a(f)$.
\STATE Observe $r_t$.
\ENDFOR
\end{algorithmic}
\end{algorithm}

\section{Main Results} \label{sec:main_results}
In this section, we provide the theoretical analysis of our algorithm. We first introduce a complexity measure for general function classes, then construct valid confidence sets, and finally derive the regret bounds in terms of Eluder Dimension.

\paragraph{Complexity Measure:} We utilize the concept of \emph{Eluder dimension}, first introduced by \cite{DBLP:conf/nips/RussoR13}. Intuitively, this measure captures the worst-case difficulty of inferring unobserved values. A low Eluder dimension implies that observing a few points allows for accurate global predictions, whereas a high dimension suggests the function can exhibit ``surprising'' variations even after many observations.
\begin{definition}[Eluder Dimension \cite{DBLP:conf/nips/RussoR13}] \label{def:eluder_dim}
An action $a$ is \emph{$\varepsilon$-independent} of a history $\{a_1, \dots, a_n\}$ with respect to $\mathcal{F}$ if there exist $f, \tilde{f} \in \mathcal{F}$ that are close on the history ($\sqrt{\sum_{i=1}^n (f(a_i) - \tilde{f}(a_i))^2} \le \varepsilon$) yet differ significantly at the new action ($|f(a) - \tilde{f}(a)| > \varepsilon$). The \emph{Eluder dimension} $\dim_E(\mathcal{F}, \varepsilon)$ is the length of the longest sequence $a_1, \dots, a_d$ such that for some $\varepsilon' \ge \varepsilon$, every element $a_k$ is $\varepsilon'$-independent of its predecessors $\{a_1, \dots, a_{k-1}\}$.
\end{definition}
To bound the regret of \texttt{NPCB}, we maintain a set $\mathcal{C}_t$ of functions
that are consistent with the observed data. The agent picks the action that
looks best under the most optimistic function in $\mathcal{C}_t$, so the
regret at each round is simply how uncertain we are about the chosen arm. We refer to this gap as the \emph{distorted width}, formally defined as $w^{\mathcal{V}}_{\mathcal{C}_t}(a) := \sup_{f \in \mathcal{C}_t}\mathcal{V}_a(f) - \inf_{f \in \mathcal{C}_t} \mathcal{V}_a(f)$. The central technical challenge is controlling this width:
since $\mathcal{V}_a(f)$ involves a non-linear distortion $\mathbb w(\cdot)$, it is not obvious that shrinking the statistical uncertainty in $f$ translates to
shrinking the subjective CPT uncertainty. We resolve this by establishing that
the CPT functional is uniformly $2$-Lipschitz in the mean reward
without any assumption on $\mathbb w(\cdot)$ (Proposition~\ref{thm:uncertainty_gap_distorted}), which immediately bounds the distorted width by the plain statistical width of $\mathcal{C}_t$ up to a constant factor (Appendix~\ref{app:technical_details},
Lemma~\ref{thm:width_bound}). This reduces CPT regret to
a standard width-shrinkage problem on $w_{\mathcal{C}_t}(a)$, to which the
Eluder dimension machinery applies directly.

\subsection{Regret for \texttt{NPCB}}

\begin{theorem}[General Regret for \texttt{NPCB}]
\label{thm:general_regret}
Under the additive reward model, with probability at
least $1-\delta$, the cumulative CPT regret $R(T)$
 satisfies

\begin{align*}
R(T)
\le \widetilde O\!\Big(
\sqrt{T\,\dim_E(\mathcal F,T^{-2})} \cdot
\sqrt{\log N(\mathcal F,T^{-2},\|\cdot\|_\infty)
+\log(1/\delta)}
\Big).
\end{align*}
\end{theorem}

\begin{remark}[Comparison with risk-neutral regret]
Theorem~\ref{thm:general_regret} has the same dependence on Eluder dimension
and covering numbers as the corresponding risk-neutral analysis.
It is sublinear when these complexities grow slowly enough with $T$.
\end{remark}

\begin{remark}
    The covering number uses the standard \emph{distance metric} sup-norm with precision $T^{-2}$.
\end{remark}
\textit{Proof Strategy:} The bound follows by invoking the cumulative width bound of
~\cite{DBLP:conf/nips/RussoR13} on the standard width, after absorbing the
CPT distortion via the $2$-Lipschitz stability of Proposition~\ref{thm:uncertainty_gap_distorted}.
The CPT factor of $2$ does not affect the leading rate.

A discussion of lower bounds and tightness is provided in
Appendix~\ref{app:lower_bounds}.

\subsection{Extension: Learning the Noise CDF Online (\texttt{NPCB-Learned})}
\label{subsec:npcb-learned}

We now drop the assumption that the arm-dependent noise CDF is known.
The mean-reward class remains $\mathcal F$, while the true noise CDF
$F$ belongs to a known class $\mathcal H$. For each $h\in\mathcal H$,
let $p_h(\cdot\mid a)$ denote its noise density at arm $a$.
Every $p_h(\cdot\mid a)$ is zero mean, supported on the common interval
$[-s,s]$, and satisfies
$0<p_{\min}\le p_h(z\mid a)\le p_{\max}$ on this support.
We also assume $|f(a)|\le B_f$ for all $f\in\mathcal F$ and $a\in\mathcal A$,
so rewards are bounded by $B_f+s$. The assumptions on the CPT weight
$\mathbb w$ are unchanged.

A candidate pair $(f,h)$ induces reward density
$q_{f,h}(r\mid a)=p_h(r-f(a)\mid a)$ and reward CDF
$G_{f,h}(z\mid a)=h(z-f(a)\mid a)$.
Let $\mathcal Q$ and $\mathcal G$ denote the resulting density and CDF
classes. Write $N_{\mathrm{opt}}(\mathcal Q,1/T)$ for the upper-density
covering number of $\mathcal Q$, and define
$\beta^{\mathcal H} =\log\frac{2N_{\mathrm{opt}}(\mathcal Q,1/T)}{\delta_0}+1 $, and
$D=\min\{\dim_E(\mathcal G,d_\cdot,1/T),T\},$
where, $d_a(G_1,G_2)=\sup_z|G_1(z\mid a)-G_2(z\mid a)|$.

At round $t$, \texttt{NPCB-Learned} keeps the usual mean confidence
set $\mathcal C_t$ and additionally forms a joint likelihood set
$\mathcal M_t\subseteq\mathcal F\times\mathcal H$. The latter retains
all pairs whose log-likelihood is at most $\beta^{\mathcal H}$ below
the maximum likelihood. The algorithm then chooses the arm with the
largest CPT value over pairs that are plausible under both confidence
sets. The full algorithm is given in
Appendix~\ref{app:learning_noise_online}.
Write
$w_{\mathcal C_t}(a)
=\sup_{f_1,f_2\in\mathcal C_t}|f_1(a)-f_2(a)|$.
\begin{theorem}[\texttt{NPCB-Learned} regret]
\label{thm:npcb-learned-regret}
Suppose $f^*\in\mathcal C_t$ for every $t\le T$ with probability at
least $1-\delta$. Then, with probability at least
$1-\delta-\delta_0$ (for a constant $C$),
\begin{align*}
R(T)
\le
C\left(1+\frac{p_{\max}}{p_{\min}}\right)
\sum_{t=1}^T w_{\mathcal C_t}(a_t) 
+\frac{C}{p_{\min}}
\left(1+D+\sqrt{D\beta^{\mathcal H}T}\right).
\label{eq:npcb-learned-regret}
\end{align*}
\end{theorem}
The first term is the usual mean-learning contribution; the second is
the additional cost of learning the noise distribution. For the
least-squares confidence sets, the first term uses the same mean-width
analysis as in the known-CDF case, with reward bound $B_f+s$.

The density constants are explicit in the bound and can be large for
sparse-tail distributions. Thus the theorem gives a uniform guarantee over the
stated class, but need not give a tight numerical prediction for
rare-event instances.

\textbf{Concrete linear example.}
Let $f_u(a)=\langle u,x(a)\rangle$, with $\|u\|_2\le B$ and
$\|x(a)\|_2\le1$, and let $\|\psi(a)\|_2\le1$. For
$v\in\mathbb R^k$, $\|v\|_2\le1$, consider
\begin{align*}
p_v(z\mid a)
&=\frac{1}{2s}
\left[
1+\langle v,\psi(a)\rangle
\left(\frac{z^2}{s^2}-\frac13\right)
\right]
\mathbf 1\{|z|\le s\}.
\end{align*}
These are zero-mean arm-dependent densities. For fixed $B,s$,
$\log N_{\mathrm{opt}}(\mathcal Q,1/T)=O((d+k)\log T)$ and
$D=O((d+k)\log T)$, giving
$R(T)=\widetilde O((d+k)\sqrt T)$.
Thus, when $k=O(d)$, learning the noise preserves the known-CDF
linear rate. This symmetric family illustrates the complexity calculation;
symmetry is not an assumption of the general theorem.
Asymmetric zero-mean noise classes are also permitted when they
satisfy the stated density and complexity conditions.
\section{Special cases: Linear \& Kernel CPT bandits}\label{sec:case_studies}

\subsection{Linear CPT Bandits}

For the linear bandit setting, $\mathcal{F} = \{\langle \cdot, \theta \rangle : \theta \in \Theta\}$
with $f_\theta(a) = \langle \theta, a \rangle$, $\theta \in \mathbb{R}^d$, where $\theta \in \mathbb R^d$ is an unknown parameter vector, and
$\mathcal{A} \subseteq \mathbb{R}^d$. Since both $\mathrm{dim}_E(\mathcal{F}, \varepsilon)$
and $\log \mathcal{N}(\mathcal{F}, \varepsilon, \|\cdot\|_\infty)$ scale as
$O(d \log(1/\varepsilon))$~\cite{DBLP:conf/nips/RussoR13}, substituting $\varepsilon = T^{-2}$
into Theorem~\ref{thm:general_regret} yields:

\begin{corollary}[Linear CPT Regret]
\label{cor:linear}
Let $\mathcal{F}$ be the unit ball of linear functions in $\mathbb{R}^d$.
Then, with probability at least $1 - \delta$,
$R(T)=\widetilde O\!\left(
d\sqrt T+\sqrt{dT\log(1/\delta)}
\right).$
\end{corollary}

\begin{remark}
For fixed confidence level, our general theorem recovers the
$\widetilde O(d\sqrt T)$ linear CPT regret order
of~\cite{DBLP:journals/corr/GopalanAFM16}.
\end{remark}

\subsection{Kernel CPT bandits}
\label{sec:kernel}

Let $\mathcal K$ be a positive semidefinite kernel on $\mathcal A$
with $\mathcal K(a,a)\le1$ for all $a$.
Let $f^*\in\mathcal H_{\mathcal K}$ with
$\|f^*\|_{\mathcal K}\le B$. The noise CDF
$F_{\epsilon_a}$ is known for every arm $a$.

For $n\ge1$ and $\rho>0$, define
$$\gamma_n(\rho)=
\sup_{a_1,\ldots,a_n\in\mathcal A}
\frac12\log\det(I+\rho^{-1}\mathbf K_n),$$
where $(\mathbf K_n)_{ij}=\mathcal K(a_i,a_j)$,
and set $\gamma_0(\rho)=0$.

\subsubsection{General \texttt{NPCB} for RKHS balls}\label{sec:kernel_specific}
We first ask what the general \texttt{NPCB} guarantee gives when
$\mathcal F$ is an RKHS ball.
A class of arbitrary continuous functions can have infinite Eluder
dimension. On any finite set of distinct arms, one can construct a
continuous function that vanishes at all previously chosen arms and
takes a nonzero value at a new arm.
Thus continuity alone does not prevent arbitrarily long independent
sequences.

An RKHS norm bound restricts this behavior.
For
$\mathcal F=\{f\in\mathcal H_{\mathcal K}:\|f\|_{\mathcal K}\le B\}$,
the Eluder dimension can be controlled through the kernel information
gain, as follows.

\begin{lemma}[Eluder dimension of an RKHS ball]
\label{thm: rkhs_finite_eluder_dim}
Let $B>0$ and
$\mathcal F=\{f\in\mathcal H_{\mathcal K}:\|f\|_{\mathcal K}\le B\}$.
For every $\varepsilon>0$, if
$\gamma_n((\varepsilon/2B)^2)\le n\log(3/2)/2$ for some $n\ge1$, then
$\dim_E(\mathcal F,\varepsilon)<n.$

Consequently, if
$\gamma_n((\varepsilon/2B)^2)=o(n)$ as $n\to\infty$, then
$\dim_E(\mathcal F,\varepsilon)<\infty$.
\end{lemma}

We now substitute these RKHS complexity bounds into the general
\texttt{NPCB} guarantee.
At $\varepsilon=T^{-2}$, Lemma~\ref{thm: rkhs_finite_eluder_dim}
evaluates information gain at the auxiliary scale
$\rho=T^{-4}/(4B^2)$.
For standard compact domains and fixed $B$, this gives
$\widetilde O(d\sqrt T)$ regret for a linear kernel and
$\widetilde O(\sqrt T(\log T)^{d+1})$ for a squared exponential
kernel.

For a Mat\'ern kernel, however, the same generic reduction is much
more conservative. Under the stated spectral conditions, \cite{vakili2021information} gives
$\dim_E(\mathcal F,T^{-2})=\widetilde O(T^{2d/\nu})$, and
\cite[Lemma~D.2]{DBLP:journals/corr/abs-2011-04622} gives
$\log N(\mathcal F,T^{-2},\|\cdot\|_\infty)
=\widetilde O(T^{2d/\nu})$.
Theorem~\ref{thm:general_regret} therefore gives at most
$\widetilde O(T^{1/2+2d/\nu})$, which may be vacuous.
This motivates exploiting the kernel structure directly.

\subsubsection{Kernel-specific confidence: \texttt{KCB}}\label{sec:improved_kernel_regret}

For an arm $a$ and scalar mean $m$, let $\mathcal V_a(m)$ denote
the distorted value of the reward $m+\epsilon_a$. Here we relax the bounded-support assumption of
Section~\ref{sec:formulation}. The noise need only be conditionally
mean zero and $\eta$-sub-Gaussian. We additionally assume that
$\int_0^\infty \mathbb w(1-F_{\epsilon_a}(z))\,dz$
and
$\int_0^\infty \mathbb w(F_{\epsilon_a}(-z))\,dz$
are finite for every arm $a$.
This ensures that CPT values are finite at every finite mean shift.
Rather than using the generic \texttt{NPCB} confidence set,
we estimate $f^*$ by kernel ridge regression.
For \texttt{KCB}, set $\lambda=1+2/T$.

At round $t$, let
$\mathbf r_{t-1}=(r_1,\ldots,r_{t-1})^\top$ and
$\mathbf k_{t-1}(a)
=(\mathcal K(a_1,a),\ldots,\mathcal K(a_{t-1},a))^\top$.
Kernel ridge regression gives the predictor $\mu_{t-1}(a)$ and
uncertainty $\sigma_{t-1}(a)$.
The algorithm forms an upper confidence bound $U_t(a)$ on the mean
and chooses the arm with the largest optimistic distorted value
$\mathcal V_a(U_t(a))$.

\begin{algorithm}[t]
\caption{Kernel CPT Bandits (\texttt{KCB})}
\label{alg:kernel_cpt}
\begin{algorithmic}[1]
\STATE \textbf{Input:} $\mathcal K,B,\eta,\delta,\lambda$ and the known noise CDFs.
\FOR{$t=1$ to $T$}
\STATE Compute $\mu_{t-1}(a)=\mathbf k_{t-1}(a)^\top
(\mathbf K_{t-1}+\lambda I)^{-1}\mathbf r_{t-1}$.
\STATE Compute $\sigma_{t-1}^2(a)=\mathcal K(a,a)-
\mathbf k_{t-1}(a)^\top(\mathbf K_{t-1}+\lambda I)^{-1}
\mathbf k_{t-1}(a)$.
\STATE Set $\sqrt{\beta_t}=B+\eta
\sqrt{2(\gamma_{t-1}(\lambda)+1+\log(1/\delta))}$.
\STATE Set $U_t(a)=\mu_{t-1}(a)+\sqrt{\beta_t}\,\sigma_{t-1}(a)$.
\STATE Choose $a_t\in\arg\max_{a\in\mathcal A}\mathcal V_a(U_t(a))$ and observe $r_t$.
\ENDFOR
\end{algorithmic}
\end{algorithm}

The estimator $\mu_{t-1}$ is the kernel ridge-regression estimator,
rather than the constrained least-squares estimator used in
Algorithm~\ref{alg:distorted-eluder}.
By \cite[Theorem~2]{DBLP:conf/icml/ChowdhuryG17}, with probability
at least $1-\delta$, simultaneously for all $t$ and $a$,
$|f^*(a)-\mu_{t-1}(a)|
\le\sqrt{\beta_t}\,\sigma_{t-1}(a)$.
Since $\mathcal V_a(m)$ is nondecreasing in $m$,
$\mathcal V_a(U_t(a))$ is an upper confidence value for the true
distorted reward of arm $a$.

\begin{theorem}[Kernel CPT regret]\label{cor: distorted_regret_RKHS}
With probability at least $1-\delta$, Algorithm~\ref{alg:kernel_cpt}
satisfies

\begin{align*}
R(T)
\le
4\sqrt{C_\lambda T\gamma_T(\lambda)} \Big[
B+\eta\sqrt{
2\big(\gamma_{T-1}(\lambda)+1+\log(1/\delta)\big)}
\Big],
\end{align*} 
where $C_\lambda=
\frac{2}{\log(1+1/\lambda)}.$
\end{theorem}

Unlike the generic \texttt{NPCB} bound, this result uses neither an
Eluder dimension nor a covering number.
For fixed $B,\eta,\delta$, it gives
$\widetilde O(d\sqrt T)$ for the linear kernel and
$\widetilde O(\sqrt T(\log T)^{d+1})$ for the squared exponential
kernel.
For Mat\'ern kernels it improves the generic bound to
$\widetilde O(T^{1/2+d/(2\nu+d)})$.

The remaining loss comes from the confidence multiplier:
$\sqrt{\beta_T}$ itself contains a
$\sqrt{\gamma_T(\lambda)}$ term, while summing the posterior
standard deviations contributes another
$\sqrt{T\gamma_T(\lambda)}$ factor.
Thus the dominant term scales as
$\sqrt T\,\gamma_T(\lambda)$.
The phasewise refinement below removes this extra
information-gain factor.

\subsubsection{Phasewise confidence refinement}
\label{sec:phasewise_cpt}

We adapt the batched pure-exploration strategy of
\cite{li2022fewbatches}. Starting from all arms, our proposed algorithm CPT Batched Pure Exploration (\texttt{CPT-BPE})
runs phases of increasing length. Within each phase, arms are selected
greedily by posterior variance using only the arms pulled in
that phase; rewards from the phase are used only after the phase ends.
A kernel confidence band is then fitted, and arms whose optimistic CPT
value is below the best pessimistic CPT value are eliminated.
The full procedure is given in
Algorithm~\ref{alg:cpt-bpe} in Appendix~\ref{app:kernel_proofs}.

\begin{theorem}[Phasewise kernel CPT regret]
\label{thm:phasewise_cpt}
Under the assumptions above, let $|r_t|\le M$ almost surely.
For finite $\mathcal A$, with probability at least $1-\delta$,
\[
R(T)\le 2M\lceil\sqrt T\rceil
+8c\sqrt{C_\lambda\gamma_T(\lambda)}
\sum_{i=2}^{m}\frac{n_i}{\sqrt{n_{i-1}}}.
\]
Here a final truncated phase uses its actual length in the sum.
In particular, this is
$\widetilde O(\sqrt{T\gamma_T(\lambda)})$
for fixed $B,\eta,M,\lambda$ and polynomial $|\mathcal A|$.
\end{theorem}
The proof is in Appendix~\ref{app:kernel_proofs}.
For a Mat\'ern kernel, the information-gain bound gives
$\widetilde O(T^{(\nu+d)/(2\nu+d)})$.
This is the risk-neutral Mat\'ern upper-bound exponent.
It holds for a finite arm set, including a finite discretization.

To extend \texttt{CPT-BPE} to a continuous arm space, let
$\mathcal A$ be a compact subset of $\mathbb R^d$ and take a grid
with mesh $T^{-1/(2\alpha)}$ for some $\alpha>0$.
Assume the true CPT objective satisfies
$|\mathcal V_a(f^*)-\mathcal V_b(f^*)|
\le L\|a-b\|^\alpha$.
Then the grid has polynomial size and introduces at most
$L\sqrt T$ additional regret, up to a domain constant.
Moreover, $|\mathcal A_T|=O(T^{d/(2\alpha)})$, so the
$\log|\mathcal A_T|$ term in the confidence multiplier contributes
only logarithmic factors, and the same Mat\'ern exponent follows.
This spatial regularity condition accounts for variation in both the
mean and the arm-dependent noise law, and is needed to compare the
continuous optimizer with its nearest grid arm.
For example, it holds if $f^*$ is $\alpha$-H\"older and
$\mathcal V_a(m)$ is $\alpha$-H\"older in $a$, uniformly over the
relevant means.
The construction and regret decomposition are given in
Appendix~\ref{app:cpt_bpe_continuous}.

\begin{table}[t!]
\centering
\begin{tabular}{@{}lccc@{}}
\toprule
\rowcolor{gray!10}
Kernel & $\gamma_T(1)$ & General \texttt{NPCB} & \texttt{CPT-BPE} \\
\midrule
Linear
& $O(d\log T)$
& $\widetilde O(d\sqrt T)$
& \cellcolor{green!10}$\widetilde O(d\sqrt T)$ \\

Sq. Exp.
& $O((\log T)^{d+1})$
& $\widetilde O(\sqrt T)$
& \cellcolor{green!10}$\widetilde O(\sqrt T)$ \\

Mat\'ern
& $\widetilde O(T^{d/(2\nu+d)})$
& \cellcolor{red!10}$\widetilde O(T^{1/2+2d/\nu})$
& \cellcolor{green!10}$\widetilde O(T^{(\nu+d)/(2\nu+d)})$ \\
\bottomrule
\end{tabular}

\caption{Comparison of known-CDF kernel regret bounds for general \texttt{NPCB} and \texttt{CPT-BPE} under standard spectral assumptions. The \texttt{CPT-BPE} column is highlighted in green. The Mat\'ern entry for general \texttt{NPCB} is highlighted in red to indicate that this bound can be vacuous.}
\label{tab:kernel_regret_comparison}
\end{table}
\subsubsection{Kernel means with learned noise CDFs}
\label{sec:kernel_learned_cdf}

Keep the $k$-parameter noise family above and replace the linear
mean class by the RKHS ball
$\mathcal F=\{f\in\mathcal H_{\mathcal K}:\|f\|_{\mathcal K}\le B\}$.
\texttt{NPCB-Learned} (Algorithm~\ref{alg:npcb-learned}) applies
unchanged.

For this RKHS specialization, the two distributional complexity terms
in Theorem~\ref{thm:npcb-learned-regret} can be controlled by the
complexity of the mean class and the $k$-dimensional noise family.
In particular,
$\log N_{\mathrm{opt}}(\mathcal Q,1/T)
\le \log N(\mathcal F,c/T,\|\cdot\|_\infty)+k\log(CT)$,
while
$\min\{\dim_E(\mathcal G,d_\cdot,1/T),T\}
\le C[\gamma_T(c_0/T^2)+k\log(1+CT^3/k)]$,
for fixed $B,s$ and constants $c,c_0,C>0$.

Together with the kernel mean-width bound
$\sum_{t=1}^T w_{\mathcal C_t}(a_t)
=O(\sqrt{T\beta_T\gamma_T(\lambda)})$,
Theorem~\ref{thm:npcb-learned-regret} gives an explicit learned-CDF
kernel guarantee. Details are given in
Appendix~\ref{kernel_with_learned_noise}.
The distributional Eluder term uses information gain at scale
$c_0/T^2$, rather than $\lambda$, so this argument does not recover
the strongest known-CDF Mat\'ern rate.

\section{Experiments}
\label{sec:experiments}

We examine how risk criteria rank arms with rare losses or gains,
how kernel algorithms learn the CPT-optimal arm, and the additional
cost of learning unknown noise laws.

\begin{figure}[!htbp]
\centering
\includegraphics[width=0.84\textwidth,height=0.22\textheight,keepaspectratio]{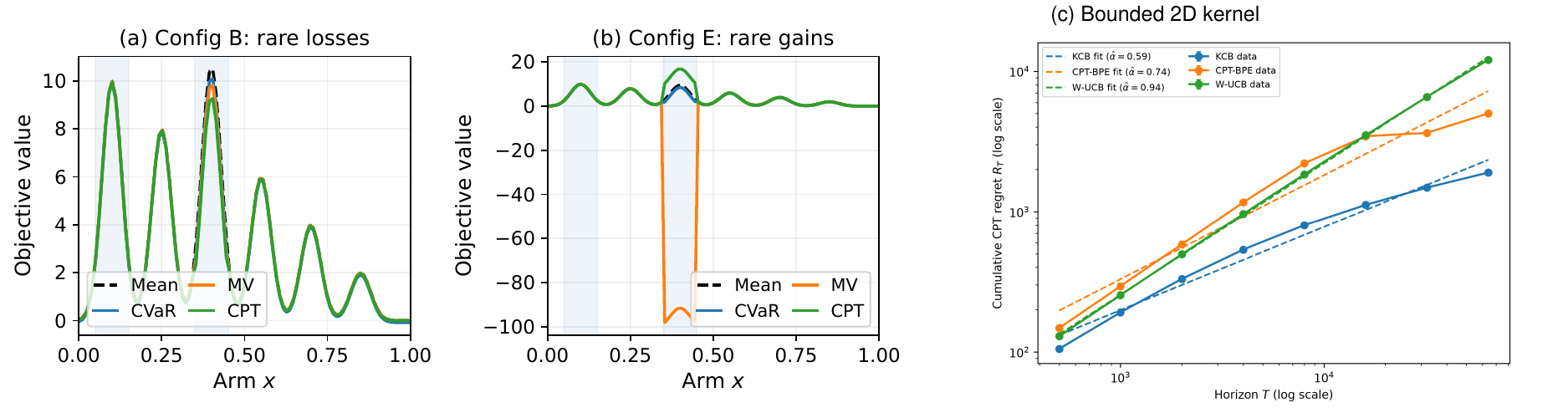}
\caption{
\textbf{Risk preferences and CPT learning.}
(a)--(b) Population objective scores in Config~B (rare loss)
and Config~E (rare gain).
CVaR uses $\alpha=0.10$, MV uses $c=1$, and CPT uses identity
utility and Tversky--Kahneman weighting with $\gamma=0.50$.
(c) Bounded 2D kernel comparison on logarithmic axes:
mean cumulative CPT regret $\pm$ one standard error over
$30$ paired trials.
Dashed lines are descriptive power-law fits.
}\label{fig:main_risk_kernel}
\end{figure}

\paragraph{Risk preferences.}
We consider $100$ candidate routes (arms), each yielding a
stochastic reward. The arms are equally spaced in $[0,1]$,
and their mean rewards form six Gaussian-shaped peaks.
We compare Mean, lower-tail CVaR, mean--variance (MV), and CPT.
Each risk metric assigns a score to every arm;
the arm with the highest score is its preferred choice.

We study five experimental configurations, labelled A--E.
Configs A--D vary the probability and magnitude of a rare loss
for arms near the third peak; config E instead introduces
a rare large gain for those arms.
The remaining arms have Gaussian background noise.
These settings illustrate contrasting human preferences
associated with small probabilities: avoiding a rare large loss
while favoring a rare large gain.

In config B, shown in Figure~\ref{fig:main_risk_kernel}(a),
arms near the third peak have higher mean rewards but carry
a small chance of a large loss.
CPT lowers their risk scores sufficiently to favor a safer arm
near the first peak, consistent with avoidance of rare severe
outcomes. MV also favors this safer alternative, whereas
Mean and CVaR retain their preference for the third peak.

Interestingly, the preference reverses for rare gains.
In config E, shown in Figure~\ref{fig:main_risk_kernel}(b),
arms near the third peak have lower mean rewards but offer
a small chance of a large gain.
CPT increases their scores sufficiently to favor them,
capturing the attraction of rare large gains.
Although MV favors the safer alternative in B, it also penalizes
the variability introduced by the rare gain in E and therefore
prefers an arm near the first peak, as do Mean and CVaR.
Thus, at the tested settings, CPT captures both contrasting
preferences within the same framework.
The remaining settings, parameter sensitivity, and learning
trajectories appear in Appendix~\ref{app:exp-part1}.

\paragraph{Kernel learning.}
We compare \texttt{KCB}, \texttt{CPT-BPE}, and empirical-CDF W-UCB
under the same CPT objective on a bounded $49$-arm grid in $[0,1]^2$.
The first two methods receive the noise CDFs and use their respective
analyzed confidence constructions; W-UCB estimates each arm's
reward CDF from samples.
We run $30$ paired trials at eight horizons from $T=500$ to $64000$.
Figure~\ref{fig:main_risk_kernel}(c) shows that \texttt{KCB}
has the lowest mean regret throughout this range.
\texttt{CPT-BPE} exhibits slower growth at larger horizons
and has lower mean regret than W-UCB from $T=16000$ onward.
Its regret per round decreases from $0.296$ to $0.078$.
Appendix~\ref{app:exp-part1} gives the confidence diagnostics,
numerical results, and additional 1D kernel and linear WOFUL comparisons.

\paragraph{Learning unknown noise laws.}
We evaluate \texttt{NPCB-Learned}
(Algorithm~\ref{alg:npcb-learned}) on a finite realizable class
containing the true mean and noise model.
Configurations A--D contain rare losses and E contains rare gains.
Table~\ref{tab:main_learned_cdf} compares it with a known-law
specialization and a greedy plug-in MLE baseline.

\texttt{NPCB-Learned} has lower mean CPT regret than plug-in MLE
in every configuration and reaches the CPT-optimal reporting
region in all $30$ trials, compared with $0$--$20\%$ for plug-in MLE.
The known-law specialization incurs substantially less regret,
demonstrating the additional cost of learning the noise law.

The finite model construction, convergence curves, and
confidence-set diagnostics appear in Appendix~\ref{app:exp-part2}.

\begin{table}[!htbp]
\centering
\caption{Unknown-CDF mean CPT regret $\pm$ s.e. over $30$ trials;
$T=(20000,10000,5000,5000,30000)$ for A--E.}
\label{tab:main_learned_cdf}

\begin{tabular}{@{}lccc@{}}
\toprule
    & \multicolumn{3}{c}{Mean CPT regret ($\pm$ s.e.)} \\
\cmidrule(lr){2-4}
Cfg & Known law & Plug-in MLE & \texttt{NPCB-Learned}\\
\midrule
A & \cellcolor{goodgreen}$35.2\pm0.1$
  & \cellcolor{badred}$2692.1\pm248.3$
  & \cellcolor{bestblue}$901.7\pm50.7$\\
B & \cellcolor{goodgreen}$33.4\pm0.0$
  & \cellcolor{badred}$5354.7\pm442.8$
  & \cellcolor{bestblue}$681.3\pm34.2$\\
C & \cellcolor{goodgreen}$33.6\pm0.1$
  & \cellcolor{badred}$1846.2\pm133.0$
  & \cellcolor{bestblue}$821.5\pm29.7$\\
D & \cellcolor{goodgreen}$33.4\pm0.0$
  & \cellcolor{badred}$2701.0\pm195.4$
  & \cellcolor{bestblue}$652.4\pm14.4$\\
E & \cellcolor{goodgreen}$50.1\pm0.0$
  & \cellcolor{badred}$204225.3\pm0.9$
  & \cellcolor{bestblue}$98458.2\pm887.7$\\
\bottomrule
\end{tabular}
\end{table}

\section{Conclusion}
We study risk-sensitive bandits beyond linear structure. For known arm-dependent noise CDFs, we derive general Eluder-dimension bounds and a pointwise kernel UCB bound; our phase-wise kernel algorithm matches the risk-neutral upper-bound exponent for finite arms, and for continuous domains under spatial regularity of the CPT objective. We further extend to learned noise CDFs via a joint likelihood approach, with explicit complexity bounds for linear and RKHS mean models. Open questions include extending CPT to RL, identifying complexity measures beyond the Eluder dimension suited to CPT, and designing implementable (approximate) algorithms for non-parametric CPT bandits.

%% file: appendix.tex
\section{Other Related Works}
Much of the recent progress in the multi-armed bandit setup has focused on adopting risk-sensitive measures like VaR and CVaR. \cite{DBLP:journals/corr/GalichetST14} introduced CVaR in the K-armed bandit setting, which was later studied extensively by \cite{Tamkin2019DistributionallyAwareEF,Tan2022CVaRRegretBF,DBLP:journals/corr/abs-1901-00997}. \cite{Baudry2020OptimalTS} established the first asymptotically optimal Thompson Sampling strategies for support-aware CVaR bandits. Several works have sought to develop unified frameworks as well. \cite{pmlr-v75-cassel18a} proposed a general framework using abstract norms, though their stability criteria are limited to Lipschitz-continuous measures and fail for singular functionals like Cumulative Prospect Theory (CPT). \cite{a2022wassersteindistanceapproachconcentration,pmlr-v258-tatli25a} address this gap under the Wasserstein metric to handle non-Lipschitz and heavy-tailed distributions. \cite{Chang2021AUT} established a Bayesian unification for ``continuous" risk functionals. \cite{liang2023distributionoptimizationframeworkconfidence} introduced a distributional optimization framework that yields tighter confidence bounds by avoiding global Lipschitz constants. Despite these advancements, all these approaches remain confined to the finite-arm setting, leaving the optimization of singular risk measures like CPT in non-parametric contextual bandits as an unaddressed challenge.

\cite{pmlr-v258-tatli25a} consider a different benchmark in which a
distortion riskmetric is applied to the reward distribution induced by
a mixture of arms.  For mixture $\pi$, their type of objective evaluates
the risk functional after forming the mixed reward distribution.
Our regret instead compares the per-round CPT value of the selected arm
with that of the best individual arm.  In particular,
$\int_{\mathcal A}\mathcal V_a(f^*)\,\pi(da)
\le
\sup_{a\in\mathcal A}\mathcal V_a(f^*),$
so randomization cannot improve our benchmark.  Applying CPT to the
mixed reward distribution
$\int_{\mathcal A}F_a^{f^*}\,\pi(da)$ is a different objective and is
not considered here.

An adjacent but distinct field is  reinforcement learning(RL), where these risk measures have also been extensively utilized to manage uncertainty in sequential decision-making. The foundation of risk-sensitive control in RL was established through entropic utility and variance-penalized objectives \cite{88583280-6268-3150-aa10-de5af371f3e0, Sobel_1982}. These were subsequently adapted into policy gradient and actor-critic algorithms to optimize exponential costs and minimize volatility by \cite{BORKAR2001339, prashanthactorcritic}. To ensure safety in high-stakes environments, research has increasingly adopted tail-risk measures like CVaR \cite{tamar2014optimizingcvarsampling, chow2017riskconstrainedreinforcementlearningpercentile}. Behavioral measures, specifically CPT, have also been integrated to align agent behavior with human decision-making and probability distortions \cite{a2016cumulativeprospecttheorymeets}.

\section{Technical Details for Section~\ref{sec:main_results}} \label{app:technical_details}
In this section, we provide the technical details supporting the regret analysis
of Section~\ref{sec:main_results}. We detail the confidence set construction,
formalize the mapping from statistical uncertainty to subjective distorted
uncertainty, and present the key intermediate results that underpin
Theorem~\ref{thm:general_regret}. The proofs of all results in this section
are provided in Appendix~\ref{app:main_results_proof}.
\subsection{Complexity Measure}
We utilize the concept of \emph{Eluder dimension}, first introduced by \cite{DBLP:conf/nips/RussoR13}. Intuitively, this measure captures the worst-case difficulty of inferring unobserved values. A low Eluder dimension implies that observing a few points allows for accurate global predictions, whereas a high dimension suggests the function can exhibit ``surprising" variations even after many observations.

\subsection{Confidence Sets}\label{sec: conf_set}

A defining characteristic of our algorithm is its modularity. Since the
statistical validity of the confidence set depends exclusively on the
additive noise properties, we can adopt the construction from
\cite{DBLP:conf/nips/RussoR13}, ensuring that the reward function's mean
concentrates objectively regardless of the agent's subjective bias.
We denote the confidence set at time $t$ as an empirical ball centered
at the least-squares estimate:
\[
  \mathcal{C}_t :=
  \left\{f \in \mathcal{F} :
  {\|f-\hat f^{LS}_t\|_{2,\mathcal D_{t-1}}}
  \le \sqrt{\beta_t}\right\}.
\]
Here, $\|g\|_{2,\mathcal D_{t-1}}^2
:=\sum_{j<t}g(a_j)^2 .$

For the confidence-set construction, it is enough that the noise is
conditionally mean zero and uniformly $\eta$-sub-Gaussian:
$
\mathbb E\!\left[
e^{u\epsilon_t}\mid \mathcal D_{t-1},a_t
\right]
\le
e^{u^2\eta^2/2},$ with $ u\in\mathbb R.$
Under the bounded-reward assumption of
Section~\ref{sec:formulation}, Hoeffding's lemma allows us to take
$\eta=M$. Moreover, every $f\in\mathcal F$ takes values in $[-M,M]$,
so the range condition of \cite{DBLP:conf/nips/RussoR13} holds with
$C=2M$.

{We set $\alpha=T^{-2}$ and assume
$N(\mathcal F,\alpha,\|\cdot\|_\infty)<\infty$.}
Following Proposition~2 of \cite{DBLP:conf/nips/RussoR13}, we use
\[
\beta_t =
8\eta^2
\log\!\left(
\frac{N(\mathcal F,\alpha,\|\cdot\|_\infty)}{\delta}
\right)
+
2\alpha t
\left(
16M+
\sqrt{8\eta^2\ln(4t^2/\delta)}
\right).
\]
This radius explicitly accounts for the function class complexity
(via $\alpha$-covering numbers in sup-norm) and the discretization
error inherent in continuous spaces.

No convexity of $\mathcal F$ is required for this
confidence construction.  This is the general-function-class setting
of \cite{DBLP:conf/nips/RussoR13}; for completeness, the following
direct argument also gives the desired $1-\delta$ confidence statement
under our standing bounded-reward assumptions.

\begin{lemma}[Least-squares confidence for general function classes]
\label{lem:ls-no-convexity}
Let $\mathcal F_\alpha\subseteq\mathcal F$ be a finite
$\alpha$-cover in the sup norm, and write
$N_\alpha=|\mathcal F_\alpha|$.
Then, with probability at least $1-\delta$, simultaneously for every
$t\le T$,
\[
\sum_{j<t}
\bigl(\hat f_t^{LS}(a_j)-f^*(a_j)\bigr)^2
\le
8\eta^2\log\frac{N_\alpha}{\delta}
+12M\alpha(t-1).
\]
In particular, with $\eta=M$, the radius $\beta_t$ above implies
$f^*\in\mathcal C_t$ for every $t\le T$.
\end{lemma}

\begin{proof}
For fixed $g\in\mathcal F_\alpha$, define
\[
Q_t(g)
:=
\sum_{j<t}
\bigl(g(a_j)-f^*(a_j)\bigr)^2,
\qquad
S_t(g)
:=
\sum_{j<t}
\epsilon_j\bigl(g(a_j)-f^*(a_j)\bigr).
\]
Conditional sub-Gaussianity implies that
\[
\exp\!\left\{
\frac{S_t(g)}{2\eta^2}
-\frac{Q_t(g)}{8\eta^2}
\right\}
\]
is a nonnegative supermartingale starting at one.
Hence, by Ville's inequality and a union bound over
$\mathcal F_\alpha$, with probability at least $1-\delta$,
simultaneously for all $g\in\mathcal F_\alpha$ and all $t\le T$,
\[
S_t(g)
\le
\frac14 Q_t(g)
+
2\eta^2\log\frac{N_\alpha}{\delta}.
\]

Work on this event and put $h=\hat f_t^{LS}$.
Least-squares optimality gives
\[
Q_t(h)\le2S_t(h).
\]
Choose $g\in\mathcal F_\alpha$ such that
$\|g-h\|_\infty\le\alpha$.
Since $r_j,f^*(a_j)\in[-M,M]$, we have
$|\epsilon_j|\le2M$, and since $g,h,f^*$ take values in $[-M,M]$,
\[
|S_t(h)-S_t(g)|
\le2M\alpha(t-1),
\qquad
|Q_t(h)-Q_t(g)|
\le4M\alpha(t-1).
\]
Therefore
\begin{align*}
Q_t(h)
&\le
\frac12Q_t(g)
+4\eta^2\log\frac{N_\alpha}{\delta}
+4M\alpha(t-1)\\
&\le
\frac12Q_t(h)
+4\eta^2\log\frac{N_\alpha}{\delta}
+6M\alpha(t-1).
\end{align*}
Rearranging proves the claim.
\end{proof}

We further define the \emph{standard width} of the confidence set at
action $a$ as the maximum variation in prediction:
\[
w_{\mathcal C_t}(a)
:=
\sup_{f_1,f_2\in\mathcal C_t}
|f_1(a)-f_2(a)|.
\]
Crucially, the Eluder dimension governs the shrinkage of this width.

\subsection{Mapping confidence set to distorted value}
While the confidence set $\mathcal{C}_t$ captures the statistical uncertainty surrounding the environment's mean reward function $f^*$ (learned from observed undistorted rewards), the agent's decisions are governed by the subjective weight-distorted value functional $\mathcal{V}_a(f)$. To bridge these, we define the optimistic and pessimistic weight-distorted rewards for any action $a$ at time $t$, respectively, as
\begin{equation*}\tilde{\mathcal{V}}_t(a) := \sup_{f \in \mathcal{C}_t} \mathcal{V}_a(f), \quad \quad \underline{\mathcal{V}}_t(a) := \inf_{f \in \mathcal{C}_t} \mathcal{V}_a(f).
\end{equation*}
Based on these boundary estimates, we define the \emph{distorted width} of the confidence set for action $a$, denoted by $w^{\mathcal{V}}_{\mathcal{C}_t}(a)$, as the subjective uncertainty gap: $w^{\mathcal{V}}_{\mathcal{C}_t}(a) := \tilde{\mathcal{V}}_t(a) - \underline{\mathcal{V}}_t(a)$.

It is important to emphasize that the confidence set $\mathcal{C}_t$ is constructed purely through square-loss estimation and is ``blind" to the specific weight distortion function $\mathbb w(\cdot)$. The distorted width $w^{\mathcal{V}}_{\mathcal{C}_t}(a)$ is introduced strictly as an analytical tool to quantify uncertainty relative to the agent's subjective preference. This decoupling ensures that our framework maintains its modularity, learn the environment using standard statistical techniques while optimizing for non-linear behavioral values.

\subsection{Regret Decomposition}

\begin{proposition}[Regret Decomposition for Distorted Rewards]
\label{prop:distorted_regret_decomp}
Assume that the true mean reward function $f^*$ is contained in the confidence set $\mathcal{C}_t$ for all rounds $t=1, \dots, T$. Then, the cumulative distorted regret $R(T)$ is (almost surely) bounded by the sum of the distorted widths of the selected actions: $R(T) \le \sum_{t=1}^T w^{\mathcal{V}}_{\mathcal{C}_t}(a_t) \text{.}$

\end{proposition}

\begin{remark} Although the agent cannot observe the true distorted rewards $\mathcal{V}_a(f^*)$, the regret bound in Proposition \ref{prop:distorted_regret_decomp} remains valid because the agent selects actions based on the optimistic estimator $\tilde{\mathcal{V}}_t$. By choosing $a_t = \arg\max_{a \in \mathcal{A}} \tilde{\mathcal{V}}_t(a)$, the algorithm guarantees that $\tilde{\mathcal{V}}_t(a_t) \ge \tilde{\mathcal{V}}_t(a^*)$. This selection rule ensures that the instantaneous regret is entirely dominated by the uncertainty, quantified as the distorted width $w^{\mathcal{V}}_{\mathcal{C}_t}(a_t)$, thereby converting the problem of minimizing unknown regret into the problem of reducing subjective uncertainty.
\end{remark}

Following Proposition \ref{prop:distorted_regret_decomp}, the theoretical core of our analysis hinges on relating the subjective \emph{distorted width} $w^{\mathcal V}_{\mathcal{C}_t}(a)$ to the \emph{standard width} $w_{\mathcal{C}_t}(a)$ of a confidence set. This modularity allows us to leverage existing concentration results for general function classes while accounting for the non-linear transformation of rewards.

\subsection{Connecting distorted width to standard width}
The confidence set $\mathcal C_t$ controls uncertainty in the mean,
whereas our regret decomposition involves the distorted width
$w^{\mathcal V}_{\mathcal C_t}(a)$.
Using the translation-stability bound
of~\cite[Lemma~5]{DBLP:journals/corr/GopalanAFM16},
we relate these widths for general mean-function classes.
This allows mean-estimation confidence guarantees to control
CPT regret.

\begin{proposition}[Uniform Lipschitz Stability of Distorted Rewards]
\label{thm:uncertainty_gap_distorted}
Assume the additive reward model, with both gain and loss CPT
integrals finite at one mean for each arm.
Let $\mathbb w:[0,1]\to[0,1]$ be Borel measurable with
$\mathbb w(0)=0$.
Then, for every arm $a$ and $f_1,f_2\in\mathcal F$,
\[
|\mathcal V_a(f_1)-\mathcal V_a(f_2)|
\le2|f_1(a)-f_2(a)|.
\]
\end{proposition}

\begin{lemma}[Bound on Distorted Width]
\label{thm:width_bound}
Under the assumptions of
Proposition~\ref{thm:uncertainty_gap_distorted}, for any action $a$,
the distorted width satisfies
$
w^{\mathcal V}_{\mathcal C_t}(a)
\le 2w_{\mathcal C_t}(a).
$
\end{lemma}

\section{Omitted Proofs from Section \ref{sec:main_results}} \label{app:main_results_proof}

\subsection{Proof of Proposition \ref{prop:distorted_regret_decomp} (Regret Decomposition for Distorted Rewards)}

\begin{proof} \label{app:proof_regret_decomp}
The instantaneous regret at round $t$ is $ir_t = \mathcal{V}_{a^*}(f_{\theta^*}) - \mathcal{V}_{a_t}(f_{\theta^*})$. Since we assume true mean reward function $f_\theta \in \mathcal{C}_t$ for all round $t$, with high probability, the true distorted reward for any action is bounded by its optimistic and pessimistic estimates: $\underline{\mathcal{V}}_t(a) \le \mathcal{V}_a(f_{\theta^*}) \le \tilde{\mathcal{V}}_t(a)$. We can therefore bound the instantaneous regret as follows:
\begin{align*}
    ir_t &= \mathcal{V}_{a^*}(f_{\theta^*}) - \mathcal{V}_{a_t}(f_{\theta^*}) \\
         &\le \tilde{\mathcal{V}}_t(a^*) - \underline{\mathcal{V}}_t(a_t) \quad (\text{since } f_{\theta^*} \in \mathcal{C}_t) \\
         &= \tilde{\mathcal{V}}_t(a_t) - \underline{\mathcal{V}}_t(a_t) + \underbrace{\left( \tilde{\mathcal{V}}_t(a^*) - \tilde{\mathcal{V}}_t(a_t) \right)}_{\le 0 \text{ by the action selection rule}} \\
         &\le \tilde{\mathcal{V}}_t(a_t) - \underline{\mathcal{V}}_t(a_t) \\
         &= w^{\mathcal{V}}_{\mathcal{C}_t}(a_t) \text{.}
\end{align*}
Summing this inequality over $t=1, \dots, T$ yields the desired result.
\end{proof}

\subsection{Proof of Proposition \ref{thm:uncertainty_gap_distorted} (Uniform Lipschitz Stability of Distorted Rewards)} \label{app:proof_lipschitz}

\begin{proof}
Fix an arm $a$ and write $V_a(m)=\mathcal V_a(m+\epsilon_a)$.
For $m_2\ge m_1$, a change of variables in the two tail integrals gives
\[
V_a(m_2)-V_a(m_1)
=\int_{-m_2}^{-m_1}
\left[\mathbb w(1-F_{\epsilon_a}(u))
+\mathbb w(F_{\epsilon_a}(u))\right]du.
\]
The integrand lies in $[0,2]$.
Therefore $0\le V_a(m_2)-V_a(m_1)\le2(m_2-m_1)$.
The same bound with the arguments exchanged proves the result.
This argument also covers noise atoms and discontinuous weight functions.
Bounded rewards ensure that both tail integrals are finite.
More generally, the same argument applies whenever both tail
integrals are finite at one mean, since they then remain finite
under every finite mean shift.

\end{proof}
\paragraph{Extension to reference points, separate weights, and Lipschitz utilities.}
Let $X_m=m+\epsilon_a$, fix a reference point $x_0$, and define
\[
U_m=u^+((X_m-x_0)^+),
\qquad
W_m=u^-((X_m-x_0)^-),
\]
where $u^\pm:[0,\infty)\to[0,\infty)$ are nondecreasing,
$L^\pm$-Lipschitz, and satisfy $u^\pm(0)=0$.
Let $\mathbb w^\pm:[0,1]\to[0,1]$ be nondecreasing with
$\mathbb w^\pm(0)=0$ and $\mathbb w^\pm(1)=1$, and set
\[
\mathcal V_a(m)
=
\int_0^\infty
\mathbb w^+\!\left(\Pr(U_m>z)\right)\,dz
-
\int_0^\infty
\mathbb w^-\!\left(\Pr(W_m>z)\right)\,dz .
\]

Couple $X_{m+\Delta}=X_m+\Delta$ for $\Delta\ge0$.
Since $x\mapsto u^+((x-x_0)^+)$ is nondecreasing and
$L^+$-Lipschitz,
\[
U_m\le U_{m+\Delta}\le U_m+L^+\Delta .
\]
Define
\[
g(z)=\mathbb w^+\!\left(\Pr(U_m>z)\right),
\qquad z\in\mathbb R.
\]
For $z<0$, $\Pr(U_m>z)=1$, while $g$ vanishes above the bounded
support of $U_m$. Hence
\[
0\le
\int_0^\infty
\left[
\mathbb w^+\!\left(\Pr(U_{m+\Delta}>z)\right)-g(z)
\right]dz
\le
\int_0^\infty [g(z-L^+\Delta)-g(z)]\,dz
\le L^+\Delta .
\]
Similarly,
\[
W_{m+\Delta}\le W_m\le W_{m+\Delta}+L^-\Delta,
\]
so the loss integral decreases by at most $L^-\Delta$. Therefore
\[
0\le
\mathcal V_a(m+\Delta)-\mathcal V_a(m)
\le
(L^++L^-)\Delta .
\]
Thus
\[
|\mathcal V_a(m_1)-\mathcal V_a(m_2)|
\le
(L^++L^-)|m_1-m_2|.
\]
Taking $x_0=0$, $u^\pm(x)=x$, and
$\mathbb w^+=\mathbb w^-=\mathbb w$ recovers the factor $2$ above.

The same CDF-sandwich argument used in
Lemma~\ref{lem:cdf_stability_new} therefore gives
\[
\left|
\mathcal V_a^{h_1}(f)-\mathcal V_a^{h_2}(f)
\right|
\le
\frac{L^++L^-}{p_{\min}}
\sup_z |h_1(z\mid a)-h_2(z\mid a)|.
\]
Consequently, the regret guarantees extend with the Lipschitz factor
$2$ replaced throughout by $L^++L^-$.

\subsection{Proof of Lemma \ref{thm:width_bound} (Bound on Distorted Width)}

\begin{proof} \label{app:proof_width_bound}
The distorted width is defined as the maximum possible difference in the distorted reward between any two functions in the confidence set:
\[
w_{\mathcal{C}_t}^\mathcal{V}(a) = \sup_{f_1, f_2 \in \mathcal{C}_t} |\mathcal{V}_a(f_1) - \mathcal{V}_a(f_2)|
\]
From Proposition \ref{thm:uncertainty_gap_distorted}, we have the pointwise bound for any pair of functions $f_1, f_2 \in \mathcal{C}_t$:
\[
|\mathcal{V}_a(f_1) - \mathcal{V}_a(f_2)| \le 2 |f_1(a) - f_2(a)|
\]
Since this inequality holds for any pair, it must also hold for the supremum over all pairs. We can therefore substitute this bound into the definition of the distorted width:
\[
w_{\mathcal{C}_t}^\mathcal{V}(a) \le 2 \cdot \sup_{f_1, f_2 \in \mathcal{C}_t} \left(  |f_1(a) - f_2(a)|\right)
\]

The term inside the parentheses is the definition of the standard width, $w_{\mathcal{C}_t}(a) = \sup_{f_1, f_2 \in \mathcal{C}_t} |f_1(a) - f_2(a)|$. Substituting this in yields the final result:
\[
w_{\mathcal{C}_t}^\mathcal{V}(a) \le 2 \left( w_{\mathcal{C}_t}(a) \right).
\]
\end{proof}

\subsection{Proof of Theorem~\ref{thm:general_regret} (General Regret for \texttt{NPCB})}
\begin{proof}
By Lemma~\ref{lem:ls-no-convexity}, with probability at least
$1-\delta$, $f^*\in\mathcal C_t$ for every $t\le T$.
On this event, Proposition~\ref{prop:distorted_regret_decomp} and
Lemma~\ref{thm:width_bound} give
$R(T)
\le 2\sum_{t=1}^T w_{\mathcal C_t}(a_t).$
The cumulative-width bound of
\cite{DBLP:conf/nips/RussoR13}, applied with accuracy $\alpha$,
confidence radius $\beta_T$, and
$w_{\mathcal C_t}(a)\le 2M$, yields
\begin{align*}
\sum_{t=1}^T w_{\mathcal C_t}(a_t)
\le T\alpha
+2M\min\{\dim_E(\mathcal F,\alpha),T\}
+4\sqrt{
T\beta_T\min\{\dim_E(\mathcal F,\alpha),T\}
}.
\end{align*}
Indeed, when $\dim_E(\mathcal F,\alpha)\le T$ this is the standard
Eluder-width bound, while otherwise the same inequality follows from
the trivial bound $w_{\mathcal C_t}(a_t)\le2M$.
Substituting $\alpha=T^{-2}$ and multiplying by $2$ gives
$R(T)\le 2T^{-1}+4M D_T+8\sqrt{T\beta_TD_T}$, where
$D_T=\min\{\dim_E(\mathcal F,T^{-2}),T\}$.
The displayed $\widetilde O$ bound follows by substituting the
definition of $\beta_T$.
\end{proof}

\section{Lower Bounds}\label{app:lower_bounds}
When $\mathbb w(p)=p$, the distorted value equals the mean reward.
Thus any risk-neutral instance that obeys the assumptions of our model
is also a CPT instance with a known noise CDF.
This gives a reduction for minimax lower bounds on that same instance class.
It does not lower-bound the realized regret of every algorithm on every instance.

\cite{scarlett2017} give the Mat\'ern minimax lower-bound exponent
$(\nu+d)/(2\nu+d)$ for a Gaussian-noise RKHS bandit model.
Their Gaussian noise has unbounded support.
Our bounded-reward assumption does not contain that model as a special case.
We therefore do not claim that this result alone proves a matching lower
bound for the bounded-reward model studied here.
Nor does our \texttt{KCB} theorem attain that exponent:
its Mat\'ern upper-bound exponent is $1/2+d/(2\nu+d)$.
The two exponents differ by $d/[2(2\nu+d)]$.
Theorem~\ref{thm:phasewise_cpt} attains the same upper-bound
exponent on its stated domain. The Gaussian-noise lower bound
still does not directly establish a lower bound for our
bounded-reward class.

\section{Kernel CPT Regret Proof}\label{app:kernel_proofs}
The algorithm uses the kernel ridge predictor from
Algorithm~\ref{alg:kernel_cpt}.
The confidence statement needed here is pointwise.
It holds for every arm and every round on one common event.

\begin{lemma}[Pointwise kernel confidence]\label{lemma:RKHS_conf_set}
Assume $\mathcal K(a,a)\le1$, $\|f^*\|_{\mathcal K}\le B$, and
conditionally mean-zero $\eta$-sub-Gaussian noise.
Use $\lambda=1+2/T$ and the definitions in
Algorithm~\ref{alg:kernel_cpt}.
With probability at least $1-\delta$, simultaneously for all $t\le T$
and $a\in\mathcal A$,
\[
|f^*(a)-\mu_{t-1}(a)|
\le\sqrt{\beta_t}\,\sigma_{t-1}(a),
\qquad
\sqrt{\beta_t}=B+\eta\sqrt{2(\gamma_{t-1}(\lambda)
+1+\log(1/\delta))}.
\]
\end{lemma}
\begin{proof}
Apply \cite[Theorem 2]{DBLP:conf/icml/ChowdhuryG17}
with its regularization $1+2/T$.
The theorem is simultaneous in the arm and round.
It applies to conditionally mean-zero sub-Gaussian noise.
\end{proof}

\begin{lemma}[Optimistic distorted value]\label{thm:optim_reduction}
On the event of Lemma~\ref{lemma:RKHS_conf_set},
\[
\mathcal V_a(f^*)\le\mathcal V_a(U_t(a))
\qquad\text{for every }a\in\mathcal A\text{ and }t\le T.
\]
\end{lemma}
\begin{proof}
The pointwise confidence bound gives $f^*(a)\le U_t(a)$.
The integral identity in the proof of
Proposition~\ref{thm:uncertainty_gap_distorted}
shows that $\mathcal V_a(m)$ is increasing in $m$.
Apply this fact to the two means.
\end{proof}

\begin{lemma}[Instantaneous CPT regret]\label{RKHS_inst_regret}
On the same event, for the arm selected by Algorithm~\ref{alg:kernel_cpt},
\[
\mathcal V_{a^*}(f^*)-\mathcal V_{a_t}(f^*)
\le4\sqrt{\beta_t}\,\sigma_{t-1}(a_t).
\]
\end{lemma}
\begin{proof}
Optimism and the choice of $a_t$ give
\[
\mathcal V_{a^*}(f^*)
\le\mathcal V_{a^*}(U_t(a^*))
\le\mathcal V_{a_t}(U_t(a_t)).
\]
The distorted value is $2$-Lipschitz in the mean.
Thus the regret is at most
$2(U_t(a_t)-f^*(a_t))$.
The two sides of the pointwise confidence bound give
$U_t(a_t)-f^*(a_t)\le2\sqrt{\beta_t}\sigma_{t-1}(a_t)$.
\end{proof}
\subsection{Proof of Theorem~\ref{cor: distorted_regret_RKHS} (Kernel CPT regret)}

\begin{proof}
Sum Lemma~\ref{RKHS_inst_regret} and apply Cauchy--Schwarz:
\[
R(T)\le4\sqrt{\beta_T}
\sqrt{T\sum_{t=1}^T\sigma_{t-1}^2(a_t)}.
\]
The determinant identity gives
\[
\frac12\log\det(I+\lambda^{-1}\mathbf K_T)
=\frac12\sum_{t=1}^T
\log(1+\sigma_{t-1}^2(a_t)/\lambda).
\]
Since $0\le\sigma_{t-1}^2(a_t)\le1$,
concavity of the logarithm gives
\[
\log(1+\sigma_{t-1}^2(a_t)/\lambda)
\ge\sigma_{t-1}^2(a_t)\log(1+1/\lambda).
\]
The selected sequence has information gain at most $\gamma_T(\lambda)$.
Therefore
\[
\sum_{t=1}^T\sigma_{t-1}^2(a_t)
\le\frac{2\gamma_T(\lambda)}{\log(1+1/\lambda)}.
\]
Substitute this bound and the definition of $\beta_T$.
\end{proof}

\begin{algorithm}[H]
\caption{CPT Batched Pure Exploration. (\texttt{CPT-BPE})}
\label{alg:cpt-bpe}
\begin{algorithmic}[1]
\STATE \textbf{Input:} finite arm set $\mathcal A$, horizon $T$,
kernel $\mathcal K$, norm bound $B$, noise bound $\eta$,
regularization $\lambda>0$, confidence level $\delta$,
weight $\mathbb w$, and known noise CDFs $\{F_{\epsilon_a}\}_{a\in\mathcal A}$.
\STATE Set $\mathcal A_1=\mathcal A$, $n_0=1$, and
$n_i=\lceil\sqrt{Tn_{i-1}}\rceil$ for $i\ge1$.
\STATE Let $m$ be the smallest integer such that
$\sum_{i=1}^{m}n_i\ge T$, and set
\[
c=
B+\frac{\eta}{\sqrt{\lambda}}
\sqrt{2\log\frac{2m|\mathcal A|}{\delta}}.
\]
\STATE Set the global round counter $t=1$.

\FOR{$i=1,\ldots,m$}
    \STATE Set the phase length
    $N_i=\min\{n_i,T-t+1\}$.
    \STATE Initialize the phase design with no sampled locations.

    \FOR{$j=1,\ldots,N_i$}
        \STATE Compute the kernel posterior variance
        $\sigma_{i,j-1}^2(a)$ using only the arms pulled earlier in phase $i$.
        \STATE Choose
        \[
        a_t\in
        \arg\max_{a\in\mathcal A_i}
        \sigma_{i,j-1}(a).
        \]
        \STATE Observe $r_t$ and store $(a_t,r_t)$ for this phase.
        \STATE Set $t\leftarrow t+1$.
    \ENDFOR

    \STATE Fit kernel ridge regression using only the rewards and
    arms from phase $i$; denote the resulting predictor and
    posterior standard deviation by $\mu_i$ and $\sigma_i$.

    \IF{$t\le T$}
        \STATE Set
        \begin{align*}
        \mathcal A_{i+1}
        =
        \Big\{a\in\mathcal A_i:\;
        \mathcal V_a\!\left(\mu_i(a)+c\sigma_i(a)\right)
        \ge
        \max_{b\in\mathcal A_i}
        \mathcal V_b\!\left(\mu_i(b)-c\sigma_i(b)\right)
        \Big\}.
        \end{align*}
    \ENDIF
\ENDFOR
\end{algorithmic}
\end{algorithm}

\begin{remark}[Fixed design within each phase]
Ties in the variance rule are broken independently of the
current-phase rewards. Conditional on the earlier phases, the arms
pulled in phase $i$ are therefore determined without using the
rewards from that phase. Hence the phase-$i$ data form the
fixed-design setting used in Lemma~\ref{lemma:offline_cpt}.
\end{remark}

\begin{lemma}[Confidence from one phase]\label{lemma:offline_cpt}
Fix a phase whose sampling locations do not use the rewards
from that phase. Conditional on the earlier phases, for each
arm $a$ and each $u>0$,
\[
\mathbb{P}\left\{|f^*(a)-\mu_i(a)|>
\left(B+\frac{\eta}{\sqrt\lambda}\sqrt{2u}\right)
\sigma_i(a)\right\}\le2e^{-u}.
\]
The phase noises may have different distributions at
different arms. They must be independent conditional on
the chosen locations and satisfy the uniform
$\eta$-sub-Gaussian bound.
\end{lemma}
\begin{proof}
Write $\phi(a)=\mathcal K(\cdot,a)$ and
$V=\lambda I+\sum_{j=1}^{n_i}\phi(a_j)\otimes\phi(a_j)$.
Then $\sigma_i^2(a)=\lambda\langle\phi(a),V^{-1}\phi(a)\rangle$.
The ridge bias is at most $B\sigma_i(a)$.
The noise term is $\sum_j b_j\epsilon_j$, where
$b_j=\langle\phi(a),V^{-1}\phi(a_j)\rangle$.
Moreover,
$\sum_j b_j^2\le \sigma_i^2(a)/\lambda$.
Hence its conditional sub-Gaussian variance proxy is at most
$\eta^2\sigma_i^2(a)/\lambda$.
Conditional sub-Gaussian concentration proves the claim.
\end{proof}

\subsection{Proof of Theorem~\ref{thm:phasewise_cpt} (Phasewise kernel CPT regret)}
\begin{proof}
Use Lemma~\ref{lemma:offline_cpt} with
$u=\log(2m|\mathcal A|/\delta)$.
A union bound gives, in every phase and for every arm,
\[
|f^*(a)-\mu_i(a)|\le c\sigma_i(a)
\]
with probability at least $1-\delta$.
The inequality also holds when earlier phases and their
surviving sets depend on past rewards: condition on that past.

Fix this event. The optimal arm $a^*$ is never eliminated.
Indeed, its upper CPT bound is at least its true value,
which is at least every lower CPT bound.
For any $a\in\mathcal A_{i+1}$, the elimination rule and
the $2$-Lipschitz property give
\[
\mathcal V_{a^*}(f^*)-\mathcal V_a(f^*)
\le4c\sigma_i(a^*)+4c\sigma_i(a)
\le8c\max_{b\in\mathcal A_i}\sigma_i(b).
\]
For clarity, each confidence interval for the CPT value
has total width at most $4c\sigma_i(a)$.

The variance-greedy choice gives
\[
n_i\max_{a\in\mathcal A_i}\sigma_i^2(a)
\le\sum_{j=1}^{n_i}\sigma_{i,j-1}^2(a_j)
\le C_\lambda\gamma_{n_i}(\lambda),
\qquad C_\lambda=\frac2{\log(1+1/\lambda)}.
\]
The first inequality holds because variance decreases as
locations are added. The second follows from the determinant
identity and $\mathcal K(a,a)\le1$.
Thus every action in phase $i+1$ has regret at most
$8c\sqrt{C_\lambda\gamma_T(\lambda)/n_i}$.
The first phase has at most $2Mn_1$ regret.
Sum these bounds over phases.
The planned lengths obey
$n_i/\sqrt{n_{i-1}}\le\sqrt T+1$.
Truncating the last phase can only decrease its contribution.
The schedule reaches the horizon in $O(\log\log T)$ phases
\citep{li2022fewbatches}. This proves the displayed bound.
\end{proof}

\subsection{From finite arms to a continuous domain}
\label{app:cpt_bpe_continuous}

Let $\mathcal A\subset\mathbb R^d$ be nonempty and compact.
Write $g(a)=\mathcal V_a(f^*(a))$ for the true CPT value.
In addition to the assumptions of Theorem~\ref{thm:phasewise_cpt},
assume $|g(a)-g(b)|\le L\|a-b\|^\alpha$ for all
$a,b\in\mathcal A$, with fixed $L<\infty$ and $\alpha>0$.
The norm is Euclidean. This assumption makes $g$ continuous,
so it attains its maximum on $\mathcal A$.

\paragraph{A finite set of feasible arms.}
For $0<h\le1$, partition a bounding box of $\mathcal A$
into cubes of side $h/\sqrt d$.
Choose one representative in $\mathcal A$ from each cube
that intersects $\mathcal A$. The resulting deterministic set
$\mathcal A_h\subset\mathcal A$ satisfies
$|\mathcal A_h|\le C_{\mathcal A}h^{-d}$,
where $C_{\mathcal A}$ depends only on the domain and $d$.
Every $a\in\mathcal A$ has a representative in
$\mathcal A_h$ at distance at most $h$.
Set $h_T=T^{-1/(2\alpha)}$ and
$\mathcal A_T=\mathcal A_{h_T}$ before collecting rewards.
Thus $|\mathcal A_T|\le C_{\mathcal A}T^{d/(2\alpha)}$.
Run Algorithm~\ref{alg:cpt-bpe} on $\mathcal A_T$,
using the original kernel and noise law at each representative.
The restriction of $f^*$ to $\mathcal A_T$ belongs to the RKHS
of the restricted kernel and has RKHS norm at most $B$.

\paragraph{Regret against the continuous optimum.}
Let $a^*\in\arg\max_{a\in\mathcal A}g(a)$ and
$a_T^*\in\arg\max_{a\in\mathcal A_T}g(a)$.
Taking a representative of $a^*$ gives
$0\le g(a^*)-g(a_T^*)\le Lh_T^\alpha$.
For the actions $a_t\in\mathcal A_T$ selected by the algorithm,
define $R_{\rm grid}(T)=\sum_{t=1}^T[g(a_T^*)-g(a_t)]$.
Then
\begin{align*}
R_{\rm cont}(T)
&=\sum_{t=1}^T[g(a^*)-g(a_t)]\\
&=R_{\rm grid}(T)+T[g(a^*)-g(a_T^*)]\\
&\le R_{\rm grid}(T)+L\sqrt T.
\end{align*}

\paragraph{Applying the finite-arm theorem.}
Use the phase schedule of Algorithm~\ref{alg:cpt-bpe} and set
\begin{align*}
c_T=B+\frac{\eta}{\sqrt\lambda}
\sqrt{2\log\frac{2m|\mathcal A_T|}{\delta}}.
\end{align*}
The confidence union bound is over the $m$ phases and
the finite set $\mathcal A_T$. No union bound over the
continuous domain is needed.
The information gain over $\mathcal A_T$ is at most
$\gamma_T(\lambda)$, since every sequence of arms in
$\mathcal A_T$ is also a sequence in $\mathcal A$.
Theorem~\ref{thm:phasewise_cpt} therefore gives, with
probability at least $1-\delta$,
\begin{align*}
R_{\rm cont}(T)
\le{}&L\sqrt T+2M\lceil\sqrt T\rceil\\
&+8c_T\sqrt{C_\lambda\gamma_T(\lambda)}
\sum_{i=2}^{m}\frac{n_i}{\sqrt{n_{i-1}}}.
\end{align*}
The final phase uses its actual length in this sum.
Polynomial grid size contributes only a logarithmic factor
to $c_T$. Hence the bound is
$\widetilde O(\sqrt{T\gamma_T(\lambda)}+\sqrt T)$
for fixed model and regularity constants.
Under the stated Mat\'ern information-gain bound, this is
$\widetilde O(T^{(\nu+d)/(2\nu+d)})$.

\paragraph{When the regularity assumption holds.}
The exponent $\alpha$ describes variation across arms,
not regularity of the probability-weighting function.
Smoothness of $f^*$ alone does not control arbitrary
arm-dependent noise laws.
A sufficient condition is
$|f^*(a)-f^*(b)|\le L_f\|a-b\|^\alpha$ and
$|\mathcal V_a(m)-\mathcal V_b(m)|
\le L_\epsilon\|a-b\|^\alpha$ uniformly over
$m\in f^*(\mathcal A)$.

For example, if the kernel satisfies
$\|\mathcal K(\cdot,a)-\mathcal K(\cdot,b)\|_{\mathcal K}
\le L_{\mathcal K}\|a-b\|^\alpha$, then the reproducing property
and $\|f^*\|_{\mathcal K}\le B$ give
$|f^*(a)-f^*(b)|
\le BL_{\mathcal K}\|a-b\|^\alpha$.
Thus one may take $L_f=BL_{\mathcal K}$.
This kernel regularity is an additional sufficient condition
and does not follow from $\mathcal K(a,a)\le1$ alone.

Using the established $2$-Lipschitz property in the mean,
\begin{align*}
|g(a)-g(b)|
\le{}&|\mathcal V_a(f^*(a))-\mathcal V_a(f^*(b))|\\
&+|\mathcal V_a(f^*(b))-\mathcal V_b(f^*(b))|\\
\le{}&(2L_f+L_\epsilon)\|a-b\|^\alpha.
\end{align*}
If all arms share the same additive noise law,
the second term vanishes and one may take $L=2L_f$.

\FloatBarrier
\section{Omitted Proofs from Section \ref{sec:case_studies}} \label{app: case_studies_proofs}

\subsection{Kernel ridge estimator}\label{app:proof_estimator_equiv}
\begin{lemma}[Regularized least squares]\label{prop: kernel_estimator_equivalence}
For every fixed $\lambda>0$, the function
\[
\mu_{t-1}=\arg\min_{f\in\mathcal H_{\mathcal K}}
\left\{\sum_{s<t}(r_s-f(a_s))^2
+\lambda\|f\|_{\mathcal K}^2\right\}
\]
satisfies
$\mu_{t-1}(a)=\mathbf k_{t-1}(a)^\top
(\mathbf K_{t-1}+\lambda I)^{-1}\mathbf r_{t-1}$.
This need not be the minimizer of unregularized square loss over
the fixed ball $\|f\|_{\mathcal K}\le B$.
\end{lemma}
\begin{proof}
The representer theorem puts the minimizer in the span of the
kernel sections at the observed arms.
Set $\boldsymbol\alpha=(\mathbf K_{t-1}+\lambda I)^{-1}
\mathbf r_{t-1}$ and
$f=\sum_{s<t}\alpha_s\mathcal K(\cdot,a_s)$.
This $f$ satisfies the normal equation for the regularized objective.
That objective is strictly convex because $\lambda>0$.
Hence $f$ is its unique minimizer, even if $\mathbf K_{t-1}$ is singular.
Evaluating $f$ at $a$ gives the formula.
\end{proof}

\subsection{Proof of Lemma \ref{thm: rkhs_finite_eluder_dim} (Eluder Dimension and Information Gain)}
\label{app:proof_eluder_dim}

\begin{proof}
Fix $\varepsilon>0$ and put
$\rho=(\varepsilon/2B)^2$.
Suppose, toward a contradiction, that
$\dim_E(\mathcal F,\varepsilon)\ge n$.
Then there are $\varepsilon'\ge\varepsilon$ and arms
$a_1,\ldots,a_n$ such that, for every $i$, some
$f_i,\tilde f_i\in\mathcal F$ satisfy
$\sum_{j<i}(f_i(a_j)-\tilde f_i(a_j))^2\le(\varepsilon')^2$
and $|f_i(a_i)-\tilde f_i(a_i)|>\varepsilon'$.

Let $g_i=f_i-\tilde f_i$. Since
$\|g_i\|_{\mathcal K}\le2B$, define
$V_{i-1}=\rho I+
\sum_{j<i}\mathcal K(\cdot,a_j)\otimes\mathcal K(\cdot,a_j)$.
Then
\begin{align*}
\|g_i\|_{V_{i-1}}^2
&=
\rho\|g_i\|_{\mathcal K}^2
+\sum_{j<i}g_i(a_j)^2
\\
&\le
\varepsilon^2+(\varepsilon')^2
\le2(\varepsilon')^2.
\end{align*}
By the reproducing property and Cauchy--Schwarz,
\begin{align*}
g_i(a_i)^2
&\le
\|g_i\|_{V_{i-1}}^2
\|\mathcal K(\cdot,a_i)\|_{V_{i-1}^{-1}}^2.
\end{align*}
Hence
$\|\mathcal K(\cdot,a_i)\|_{V_{i-1}^{-1}}^2>1/2$.

The determinant identity therefore gives
\begin{align*}
\log\det(I+\rho^{-1}\mathbf K_n)
&=
\sum_{i=1}^n
\log\!\left(
1+\|\mathcal K(\cdot,a_i)\|_{V_{i-1}^{-1}}^2
\right)
\\
&>
n\log(3/2).
\end{align*}
For the particular sequence $a_1,\ldots,a_n$ constructed above,
\begin{align*}
\frac12\log\det(I+\rho^{-1}\mathbf K_n)
>
\frac n2\log(3/2).
\end{align*}
Since $\gamma_n(\rho)$ is the supremum of this quantity over all
sequences of $n$ arms,
$\gamma_n(\rho)>n\log(3/2)/2$.
This contradicts the assumption
$\gamma_n(\rho)\le n\log(3/2)/2$.
Therefore $\dim_E(\mathcal F,\varepsilon)<n$.

If $\gamma_n(\rho)=o(n)$, there exists an integer $N$ such that
$\gamma_N(\rho)\le N\log(3/2)/2$.
The result just proved then yields
$\dim_E(\mathcal F,\varepsilon)<N<\infty$.
\end{proof}

This comparison follows from
\cite[Theorem~3.1]{DBLP:journals/corr/abs-2107-02377},
with the factor $1/2$ accounting for our definition of $\gamma_n$.

For completeness, put $p=d/(2\nu+d)$.
Under the Mat\'ern spectral conditions of
\cite[Theorem~3 and Corollary~1]{vakili2021information}, their
information-gain estimate, with explicit dependence on the positive
regularization scale $\lambda$, gives
$\gamma_n(\lambda)=\widetilde O((n/\lambda)^p)$,
where $p=d/(2\nu+d)$.
Choose $n$ of order
$\lambda^{-p/(1-p)}$ up to logarithmic factors.
The information gain is then at most
$n\log(3/2)/2$ after increasing the constant.
Since $p/(1-p)=d/(2\nu)$ and
$\lambda=(\varepsilon/2B)^2$, the lemma gives
$\dim_E(\mathcal F,\varepsilon)
=\widetilde O((B/\varepsilon)^{d/\nu})$.
At $\varepsilon=T^{-2}$ this is
$\widetilde O(T^{2d/\nu})$ for fixed $B$.

\section{The CPT Weight Function: Behavioral Interpretation and Examples}
\label{app:weight_function}

The weight function $\mathbb w:[0,1]\to[0,1]$ governs the physical meaning of the
distorted value $V_a(f)$. The key behavioral properties required to capture
human decision-making are: $\mathbb w(0)=0$, $\mathbb w(1)=1$ (boundary conditions); $\mathbb w(p) >
p$ for small $p$ (overweighting of rare events); $\mathbb w(p) < p$ for large $p$
(underweighting of common events); monotone increasing; and an inverse S-shape
with inflection near the endpoints. Our framework requires only that $\mathbb w$ is
monotone increasing with $\mathbb w(0)=0$, $\mathbb w(1)=1$ --- all other properties are
behavioral desiderata, not mathematical requirements.

The choice of $\mathbb w$ directly governs the physical meaning of $V_a(f)$. When $\mathbb w$
is the identity, $V_a(f)$ reduces to the standard expectation (risk-neutral).
An S-shaped $\mathbb w$ penalizes heavy tails and rare catastrophic losses, favoring
safe arms over high-mean-but-risky ones. A more distorted $\mathbb w$ moves $V_a(f)$
further from the mean, potentially reversing the preference order entirely.
Several families satisfying these properties are widely used in practice:

\begin{itemize}
    \item \cite{tversky1992advances}: $\mathbb w(p) = p^\gamma/(p^\gamma + (1-p)^\gamma)^{1/\gamma}$
    \item \cite{prelec1998probability}: $\mathbb w(p) = \exp(-(-\ln p)^\alpha)$
    \item \cite{gonzalez1999shape}: $\mathbb w(p) = \delta p^\gamma/(\delta p^\gamma + (1-p)^\gamma)$
   \item Neo-additive~\cite{chateauneuf2007choice}: for $p\in(0,1)$,
      $\mathbb w(p)=\delta_0+(1-\delta_0-\delta_1)p$, with
      $\mathbb w(0)=0$ and $\mathbb w(1)=1$, where
      $\delta_0,\delta_1\ge0$ and $\delta_0+\delta_1\le1$.
      It has endpoint jumps whenever $\delta_0>0$ or $\delta_1>0$.
\end{itemize}

\paragraph{On removing H\"{o}lder continuity.}
Prior bandit work~\cite{DBLP:journals/corr/GopalanAFM16} assumes $\mathbb w$ to be $\alpha$-H\"{o}lder
continuous. This assumption
is a mathematical convenience, not a behavioral necessity. Kahneman and Tversky explicitly discuss ``apparent discontinuities
at the endpoints'' in their original prospect-theory treatment
\cite{kahneman1979prospect}. Neo-additive capacities provide an
explicit discontinuous example: for $p\in(0,1)$,
$w(p)=\delta_0+(1-\delta_0-\delta_1)p$, with $w(0)=0$ and $w(1)=1$.
Such endpoint jumps violate every H\"{o}lder condition on $[0,1]$.
Proposition~\ref{thm:uncertainty_gap_distorted} and our learned-CDF
analysis do not require H\"{o}lder continuity; monotonicity together
with the endpoint conditions is sufficient.

\section{Learning the noise online: CPT bandits without a known noise CDF}
\label{app:learning_noise_online}

The algorithm \ref{alg:distorted-eluder} in the main paper uses the noise CDF at each arm. Here we learn
that CDF from the observed rewards. We keep the additive model
$r_t=f^*(a_t)+\epsilon_t$. Conditional on the past and the chosen arm,
$\epsilon_t$ has the specified arm-dependent law. The actions may be chosen
adaptively.

\paragraph{Assumptions.}
Let $\mathcal H$ be a known class of conditional noise CDFs, with true member
$F$. For each $h\in\mathcal H$, let $p_h$ be its density and write
$h(z\mid a)=\int_{-s}^{z}p_h(u\mid a)\,du$, extended by $0$ and $1$ outside
$S=[-s,s]$. We assume that every $p_h(\cdot\mid a)$ is supported on $S$, has
mean zero, and satisfies
\[
0<p_{\min}\le p_h(z\mid a)\le p_{\max}<\infty,
\qquad z\in S,\quad a\in\mathcal A.
\]
We also assume $\sup_{f\in\mathcal F,a\in\mathcal A}|f(a)|\le B_f$ and
$(f^*,F)\in\mathcal F\times\mathcal H$. Thus
$|r_t|\le B_f+s$, and the noise is $s$-sub-Gaussian. For the mean confidence
set $\mathcal C_t$, we may use the full-history set of
Section~\ref{sec: conf_set}, with its radius $\beta_t$ chosen using these
bounds. More generally, we choose $\mathcal C_t$ so that
$f^*\in\mathcal C_t$ for every $t\le T$ with probability at least
$1-\delta$. We assume the weight
$\mathbb w:[0,1]\to[0,1]$ is nondecreasing, with
$\mathbb w(0)=0$ and $\mathbb w(1)=1$. No continuity is required.

For $(f,h)\in\mathcal F\times\mathcal H$, define the conditional reward
density and CDF by
\[
q_{f,h}(r\mid a)=p_h(r-f(a)\mid a),\qquad
G_{f,h}(z\mid a)=h(z-f(a)\mid a).
\]
Let $\mathcal Q$ and $\mathcal G$ be the resulting density and CDF classes.
We use an \emph{upper density cover} of $\mathcal Q$. At precision
$\varepsilon>0$, it is a finite set of nonnegative measurable functions
$\overline{\mathcal Q}$ such that, for every $q\in\mathcal Q$, some
$\bar q\in\overline{\mathcal Q}$ satisfies, for all $a,r$,
\[
\bar q(r\mid a)\ge q(r\mid a),\qquad
\int\bigl(\bar q(r\mid a)-q(r\mid a)\bigr)\,dr\le\varepsilon.
\]
The cover functions need not integrate to one. Write
$N_{\mathrm{opt}}(\mathcal Q,\varepsilon)$ for the smallest cover size, and
assume $N_{\mathrm{opt}}(\mathcal Q,1/T)<\infty$. This is a condition on the
\emph{shifted reward densities}. A cover of the unshifted CDFs alone does not
control their likelihoods when the support moves with $f(a)$.

For $G_1,G_2\in\mathcal G$, put
$d_a(G_1,G_2)=\sup_z|G_1(z\mid a)-G_2(z\mid a)|$.
We use $\dim_E(\mathcal G,d_\cdot,\varepsilon)$ as in
Definition~\ref{def:eluder_dim}, with the scalar difference replaced by
$d_a$. In particular, a new action is $\varepsilon$-independent if two
members of $\mathcal G$ differ there by more than some
$\varepsilon'\ge\varepsilon$, while the sum of their squared distances at
earlier actions is at most $(\varepsilon')^2$.

\paragraph{Algorithm.}
Let $\delta_0\in(0,1)$, and set
\[
\beta^{\mathcal H}
=\log\!\frac{2N_{\mathrm{opt}}(\mathcal Q,1/T)}{\delta_0}+1.
\]
Using the raw rewards, form the likelihood set
\[
\mathcal M_t=\left\{(f,h)\in\mathcal F\times\mathcal H:
\sum_{\tau<t}\log q_{f,h}(r_\tau\mid a_\tau)
\ge \sup_{(f',h')\in\mathcal F\times\mathcal H}
\sum_{\tau<t}\log q_{f',h'}(r_\tau\mid a_\tau)
-\beta^{\mathcal H}\right\}.
\]
At round $t$, choose an arm maximizing
\[
\widetilde{\mathcal V}_t(a)
=\sup_{(f,h)\in\mathcal M_t,\,f\in\mathcal C_t}
\mathcal V_a^h(f).
\]
Here $\mathcal V_a^h(f)$ is the CPT value defined in the main paper, using
$h(z-f(a)\mid a)$ as the reward CDF. We assume the displayed suprema are
attained. The analysis also permits approximate maximizers, with their
optimization errors added to the regret. See, Algorithm \ref{alg:npcb-learned} for the pseudocode.

\begin{algorithm}[t]
\caption{\texttt{NPCB-Learned}}
\label{alg:npcb-learned}
\begin{algorithmic}[1]
\STATE \textbf{Input:}
$\mathcal F$, $\mathcal H$, confidence schedules
$\{\beta_t\}$ and $\beta^{\mathcal H}$.
\FOR{$t=1,\ldots,T$}
    \STATE Construct the mean confidence set $\mathcal C_t$
    from all observations before round $t$.
    \STATE For $(f,h)\in\mathcal F\times\mathcal H$, compute
    \begin{align*}
    \ell_t(f,h)
    &=
    \sum_{\tau<t}
    \log p_h(r_\tau-f(a_\tau)\mid a_\tau).
    \end{align*}
    \STATE Form the likelihood confidence set
    \begin{align*}
    \mathcal M_t
    =
    \Big\{(f,h):
    \ell_t(f,h)
    \ge
    \sup_{(f',h')\in\mathcal F\times\mathcal H}
    \ell_t(f',h')
    -\beta^{\mathcal H}
    \Big\}.
    \end{align*}
    \STATE Choose
    $
    a_t\in\arg\max_{a\in\mathcal A}
    \sup_{\substack{(f,h)\in\mathcal M_t\\f\in\mathcal C_t}}
    \mathcal V_a^h(f).
  $
    \STATE Observe $r_t$.
\ENDFOR
\end{algorithmic}
\end{algorithm}

\begin{lemma}[Likelihood confidence set]
\label{lem:learned_cdf_likelihood}
With probability at least $1-\delta_0$, simultaneously for all $t\le T$,
$(f^*,F)\in\mathcal M_t$. On the same event, every
$(f,h)\in\mathcal M_t$ satisfies
\[
\sum_{\tau<t}d_{a_\tau}(G_{f,h},G_{f^*,F})^2
\le C\beta^{\mathcal H}
\]
for a numerical constant $C$.
\end{lemma}

\begin{proof}
Write $q^*=q_{f^*,F}$, $N=N_{\mathrm{opt}}(\mathcal Q,1/T)$,
$L=\log(2N/\delta_0)$, and $\beta^{\mathcal H}=L+1$.
For each upper cover member $\bar q$, define
$Z_{\bar q}(a)=\int\bar q(r\mid a)\,dr$ and
$\widetilde q(r\mid a)=\bar q(r\mid a)/Z_{\bar q}(a)$.
Then $1\le Z_{\bar q}(a)\le1+1/T$.
Under the true law, the likelihood-ratio process
\[
\Lambda_t(\widetilde q)=
\prod_{j<t}\frac{\widetilde q(r_j\mid a_j)}{q^*(r_j\mid a_j)}
\]
is a nonnegative supermartingale. Its conditional expected multiplier is
$\int_{\{q^*>0\}}\widetilde q(r\mid a_t)\,dr\le1$;
the inequality may be strict when $\widetilde q$ places mass outside
the support of $q^*$.
Ville's inequality and a union bound imply
$\log\Lambda_t(\widetilde q)\le L$ for all cover members and all
$t\le T$, except on an event of probability $\delta_0/2$.
If $q\le\bar q$, then
\[
\ell_t(q)-\ell_t(q^*)
\le\log\Lambda_t(\widetilde q)
   +\sum_{j<t}\log Z_{\bar q}(a_j)
\le L+1=\beta^{\mathcal H}.
\]
Hence the true pair belongs to $\mathcal M_t$.

Define $H^2(p,q)=\int(\sqrt p-\sqrt q)^2$ and
$H_a^2=H^2(\widetilde q(\cdot\mid a),q^*(\cdot\mid a))$.
The process
\[
\Lambda_t(\widetilde q)^{1/2}
\exp\left\{\frac12\sum_{j<t}H_{a_j}^2\right\}
\]
is also a nonnegative supermartingale, because its one-step
conditional expectation is
$e^{H_a^2/2}(1-H_a^2/2)\le1$.
A second maximal inequality, with failure probability $\delta_0/2$,
gives simultaneously
\[
\tfrac12\log\Lambda_t(\widetilde q)
+\tfrac12\sum_{j<t}H_{a_j}^2\le L.
\]
For $q\in\mathcal M_t$, its normalized upper cover satisfies
$\log\Lambda_t(\widetilde q)\ge-\beta^{\mathcal H}-1$.
Thus $\sum_{j<t}H_{a_j}^2\le3\beta^{\mathcal H}$.
Furthermore,
$\operatorname{TV}(q(\cdot\mid a),\widetilde q(\cdot\mid a))\le1/T$.
Using $H^2\le2\operatorname{TV}$ and the Hellinger triangle inequality,
\[
\sum_{j<t}H^2(q(\cdot\mid a_j),q^*(\cdot\mid a_j))
\le6\beta^{\mathcal H}+4\le10\beta^{\mathcal H}.
\]
Finally, Kolmogorov distance is at most total variation, which is
at most $H$ under this convention. The claimed bound follows.
All conditional arguments allow adaptively selected arms.
\end{proof}

\begin{lemma}[Stability without a smooth weight]
\label{lem:cdf_stability_new}
For any $f,a$ and $h_1,h_2\in\mathcal H$,
\[
|\mathcal V_a^{h_1}(f)-\mathcal V_a^{h_2}(f)|
\le\frac{2}{p_{\min}}\sup_z|h_1(z\mid a)-h_2(z\mid a)|.
\]
For a fixed noise CDF $h$, the value is nondecreasing and $2$-Lipschitz in
the mean $f(a)$.
\end{lemma}
\begin{proof}
Write $m=f(a)$. If $m_1\ge m_2$, a change of variables in the two CPT
integrals gives
\[
\mathcal V_a^h(m_1)-\mathcal V_a^h(m_2)
=\int_{-m_1}^{-m_2}
\bigl\{\mathbb w(1-h(u\mid a))+\mathbb w(h(u\mid a))\bigr\}\,du.
\]
The integrand belongs to $[0,2]$. This proves the claim about the mean,
including for discontinuous weights.

Now let $\xi=\sup_z|h_1(z\mid a)-h_2(z\mid a)|$ and
$\Delta=\xi/p_{\min}$. The common support and the density lower bound give
\[
h_1(z-\Delta\mid a)\le h_2(z\mid a)
\le h_1(z+\Delta\mid a),\qquad z\in\mathbb R.
\]
At either end of the support, the relevant CDF is $0$ or $1$. Pointwise CDF
order reverses CPT value order because $\mathbb w$ is nondecreasing.
The preceding mean bound therefore sandwiches
$\mathcal V_a^{h_2}(f)$ between
$\mathcal V_a^{h_1}(m-\Delta)$ and
$\mathcal V_a^{h_1}(m+\Delta)$. The result follows.
\end{proof}

\begin{lemma}[Stability for Lipschitz weights]
\label{lem:cdf_stability_lipschitz_weight}
Suppose $\mathbb w$ is $L_{\mathbb w}$-Lipschitz.
Let $h_1(\cdot\mid a)$ and $h_2(\cdot\mid a)$ be noise CDFs
supported on $[-s,s]$, and let
$\xi=\sup_z|h_1(z\mid a)-h_2(z\mid a)|$.
For every fixed mean $m$,
$|\mathcal V_a^{h_1}(m)-\mathcal V_a^{h_2}(m)|
\le 2sL_{\mathbb w}\xi$.
No density assumption is needed.
\end{lemma}

\begin{remark}
Lemma~\ref{lem:cdf_stability_lipschitz_weight} permits atomic noise
laws but requires Lipschitz weights. Lemma~\ref{lem:cdf_stability_new} instead permits
discontinuous weights under the stated density assumptions.
These are alternative sufficient conditions for stability.
The Lipschitz-weight lemma alone does not establish a learned-CDF
regret theorem for atomic models; their likelihood confidence
sets and distributional complexities require separate analysis.
\end{remark}

\begin{proof}
Apply the Lipschitz inequality inside the gain and loss integrals.
After changing variables, their domains partition the real line,
up to the single point $-m$. Hence the sum is bounded by
$L_{\mathbb w}\int_{\mathbb R}
|h_1(u\mid a)-h_2(u\mid a)|\,du$.
The CDFs agree outside $[-s,s]$, giving the claimed bound.
\end{proof}

\begin{theorem}[Regret with a learned noise CDF]
\label{thm:main}
Let $w_{\mathcal C_t}(a)=\sup_{f_1,f_2\in\mathcal C_t}
|f_1(a)-f_2(a)|$, and put
\[
D=\min\{\dim_E(\mathcal G,d_\cdot,1/T),T\}.
\]
Under the assumptions above, with probability at least
$1-\delta-\delta_0$, the algorithm satisfies
\[
R(T)\le C\left(1+\frac{p_{\max}}{p_{\min}}\right)
\sum_{t=1}^T w_{\mathcal C_t}(a_t)
+\frac{C}{p_{\min}}
\left(1+D+\sqrt{D\beta^{\mathcal H}T}\right).
\]
For the least-squares sets, the first sum has the mean-width bound of
\cite[Lemma~2]{DBLP:conf/nips/RussoR13}, using the new reward bound
$B_f+s$. No bound on $D$ follows from the mean-function complexity alone.
\end{theorem}

\begin{proof}
Work on the intersection of the two confidence events, which holds
with probability at least $1-\delta-\delta_0$. Thus
$f^*\in\mathcal C_t$ and $(f^*,F)\in\mathcal M_t$ for every
$t\le T$.

By optimism and the choice of $a_t$,
\begin{align*}
\mathcal V^F_{a^*}(f^*)-\mathcal V^F_{a_t}(f^*)
&\le
\sup_{\substack{(f,h)\in\mathcal M_t\\ f\in\mathcal C_t}}
\left\{
\mathcal V^h_{a_t}(f)-\mathcal V^F_{a_t}(f^*)
\right\}.
\end{align*}
Lemma~H.2 therefore gives
\begin{align*}
\mathcal V^F_{a^*}(f^*)-\mathcal V^F_{a_t}(f^*)
\le
2w_{\mathcal C_t}(a_t)
+\frac{2}{p_{\min}}
\sup_{\substack{(f,h)\in\mathcal M_t\\ f\in\mathcal C_t}}
\sup_z|h(z\mid a_t)-F(z\mid a_t)|.
\end{align*}
For every $(f,h)\in\mathcal M_t$ with $f\in\mathcal C_t$,
the density upper bound and $f^*\in\mathcal C_t$ imply
\begin{align*}
\sup_z|h(z\mid a_t)-F(z\mid a_t)|
&\le
d_{a_t}(G_{f,h},G_{f^*,F})
+p_{\max}w_{\mathcal C_t}(a_t).
\end{align*}
Hence
\begin{align*}
R(T)
\le
2\left(1+\frac{p_{\max}}{p_{\min}}\right)
\sum_{t=1}^T w_{\mathcal C_t}(a_t)
+\frac{2}{p_{\min}}
\sum_{t=1}^T
\sup_{(f,h)\in\mathcal M_t}
d_{a_t}(G_{f,h},G_{f^*,F}).
\end{align*}

It remains to bound the second sum. By Lemma~H.1, every
$(f,h)\in\mathcal M_t$ satisfies
\begin{align*}
\sum_{j<t}
d_{a_j}(G_{f,h},G_{f^*,F})^2
\le 10\beta^{\mathcal H}.
\end{align*}
Fix $\epsilon\ge1/T$. Whenever
\begin{align*}
\sup_{(f,h)\in\mathcal M_t}
d_{a_t}(G_{f,h},G_{f^*,F})>\epsilon,
\end{align*}
choose a witnessing pair $(f,h)\in\mathcal M_t$.
The preceding historical bound and the standard
independent-sequence counting argument give
\begin{align*}
\#\left\{
t:
\sup_{(f,h)\in\mathcal M_t}
d_{a_t}(G_{f,h},G_{f^*,F})>\epsilon
\right\}
\le
D\left(1+\frac{10\beta^{\mathcal H}}{\epsilon^2}\right).
\end{align*}
Indeed, the relevant rounds can be partitioned into at most
$1+10\beta^{\mathcal H}/\epsilon^2$ subsequences on each of which
successive actions are $\epsilon$-independent; each such subsequence
has length at most $D$.

Since $d_a$ is the Kolmogorov distance between two CDFs,
$d_a(G_1,G_2)\le1$. Integrating the preceding counting bound over
$\epsilon\in[1/T,1]$, together with the trivial count $T$ below the
optimizing threshold, yields
\begin{align*}
\sum_{t=1}^T
\sup_{(f,h)\in\mathcal M_t}
d_{a_t}(G_{f,h},G_{f^*,F})
\le
C\left(
1+D+\sqrt{D\beta^{\mathcal H}T}
\right)
\end{align*}
for a numerical constant $C$. If the optimizing threshold exceeds
one, the same conclusion follows from the trivial bound by $T$.

Substituting this estimate above gives
\begin{align*}
R(T)
&\le
C\left(1+\frac{p_{\max}}{p_{\min}}\right)
\sum_{t=1}^T w_{\mathcal C_t}(a_t)
\\[-0.2em]
&\quad+
\frac{C}{p_{\min}}
\left(
1+D+\sqrt{D\beta^{\mathcal H}T}
\right),
\end{align*}
which proves the claim.
\end{proof}

\begin{remark}[Density constants and rare events]
If a tail region of width $w_\tau$ has mass $p$ and a bulk region
of width $w_b$ has mass $1-p$, then
$p_{\min}\le p/w_\tau$ and $p_{\max}\ge(1-p)/w_b$.
Hence
$p_{\max}/p_{\min}\ge(1-p)w_\tau/(pw_b)$.
For fixed positive widths, this ratio is at least of order $1/p$;
it can be larger. The bound can therefore be conservative for
sparse-tail laws. We do not claim that this density dependence is
minimax optimal. The expected waiting time $1/p$ concerns pulls of the relevant arm,
not global rounds. It does not alone establish a regret lower bound:
non-tail observations or shared structure may also distinguish
candidate noise laws.
\end{remark}
\begin{remark}
    The factor $p_{\max}/p_{\min}$ arises when converting uncertainty
in the shifted reward CDF into uncertainty in the noise CDF.
The density upper bound controls the uncertain mean shift, and
the density lower bound converts CDF error into CPT-value error.
We do not claim that this factor is necessary.
\end{remark}

\begin{remark}[Kernel mean confidence sets]
In the RKHS case, one may instead use the pointwise confidence band
\[
\mathcal C_t=\left\{f\in\mathcal F:
|f(a)-\mu_{t-1}(a)|\le\sqrt{\beta_t}\,\sigma_{t-1}(a)
\ \text{for every }a\in\mathcal A\right\}.
\]
With $\beta_t$ and the regularization chosen as in the uniform kernel
confidence theorem of \cite{DBLP:conf/icml/ChowdhuryG17}, it contains
$f^*$ with probability at least $1-\delta$. Its width is at most
$2\sqrt{\beta_t}\,\sigma_{t-1}(a)$, so its contribution to the regret bound
is $O((1+p_{\max}/p_{\min})\sqrt{T\beta_T\gamma_T})$.
The learned-CDF term still depends on $D$ and
$N_{\mathrm{opt}}(\mathcal Q,1/T)$. These complexities need separate bounds
for a given noise model.
\end{remark}

\begin{lemma}[An arm-dependent family with explicit complexity]
\label{lem:learned_cdf_example}
Suppose $f_u(a)=\langle u,x(a)\rangle$, where $u\in\mathbb R^d$,
$\|u\|_2\le B$, and $\|x(a)\|_2\le1$. Let $\psi(a)\in\mathbb R^k$ satisfy
$\|\psi(a)\|_2\le1$. For $\|v\|_2\le1$, put
$\theta_v(a)=\langle v,\psi(a)\rangle$ and
\[
p_v(z\mid a)=\frac{1}{2s}
\left(1+\theta_v(a)\left(\frac{z^2}{s^2}-\frac13\right)\right)
\mathbf 1\{|z|\le s\}.
\]
This is a zero-mean, arm-dependent density. It satisfies
$p_{\min}=1/(6s)$ and $p_{\max}=5/(6s)$. For fixed $B,s$ and
$T\ge2$,
\[
\log N_{\mathrm{opt}}(\mathcal Q,1/T)
=O((d+k)\log T),\qquad
\min\{\dim_E(\mathcal G,d_\cdot,1/T),T\}
=O((d+k)\log T).
\]
Consequently, the theorem gives
$R(T)=\widetilde O((d+k)\sqrt T)$, with constants depending on
$B,s$ and the confidence levels.
\end{lemma}
\begin{proof}
The densities have uniformly bounded variation, including their jumps
at $-s$ and $s$. A uniform grid for $u$ and $v$, followed by a pointwise
upper envelope over each grid cell, gives an upper density cover.
Moving $u$ by $\rho$ changes the location by at most $\rho$.
The extra mass in its envelope is $O(\rho)$ by bounded variation.
Moving $v$ by $\rho$ adds $O(\rho)$ mass as well.
Taking $\rho$ of order $1/T$ gives a cover of size at most
$(CT)^{d+k}$.

For the Eluder bound, let $m=f_u(a)$ and $\theta=\theta_v(a)$.
All reward laws are supported on $[-B-s,B+s]$ and have mean $m$.
Thus a Kolmogorov gap $\xi$ between two reward laws implies
$|m-m'|\le2(B+s)\xi$. At $z=s/2$, the noise CDFs differ by
$|\theta-\theta'|/16$. Shifting their arguments by $m-m'$ changes
a CDF by at most $p_{\max}|m-m'|$. Hence, for constants $c,C>0$,
\[
c\sqrt{(m-m')^2+(\theta-\theta')^2}
\le d_a(G_{u,v},G_{u',v'})
\le C\sqrt{(m-m')^2+(\theta-\theta')^2}.
\]
The right side is a two-coordinate linear prediction error in the
parameter $(u,v)\in\mathbb R^{d+k}$. For a sequence independent at
scale $\varepsilon'\ge1/T$, apply the determinant argument to
$\lambda I+\sum_{\tau<t}\operatorname{diag}
(x(a_\tau)x(a_\tau)^\top,\psi(a_\tau)\psi(a_\tau)^\top)$,
with $\lambda$ proportional to $(\varepsilon')^2$.
Each independent action increases its log determinant by a fixed
positive amount. Its total increase over at most $T$ actions is
$O((d+k)\log T)$. This proves the second bound.
The linear mean-confidence width has sum
$\widetilde O(d\sqrt T)$ by
\cite[Lemma~2]{DBLP:conf/nips/RussoR13}. Substitution proves the
regret claim. When $k=O(d)$, the rate is $\widetilde O(d\sqrt T)$.
Thus this family retains the known-CDF linear horizon rate.
\end{proof}

The complexity bounds above apply to this particular family.
We do not assert that they hold for arbitrary skewed families.
The experimental finite noise class is a separate instantiation
and is not justified by this lemma.

\subsection{Kernel means with learned noise CDFs} \label{kernel_with_learned_noise}
\begin{lemma}[Kernel means with the explicit noise family]
\label{lem:learned_cdf_kernel_example}
Replace the linear mean in Lemma~\ref{lem:learned_cdf_example} by
$\mathcal F=\{f\in\mathcal H_{\mathcal K}:\|f\|_{\mathcal K}\le B\}$.
Assume $\mathcal K(a,a)\le1$ and $k\ge1$.
Assume the sup-norm cover below is finite. Keep its noise densities $p_v$ and
$\|\psi(a)\|_2\le1$. For constants depending only on $B$ and $s$,
\[
\log N_{\mathrm{opt}}(\mathcal Q,1/T)
\le \log N(\mathcal F,c/T,\|\cdot\|_\infty)
   +k\log(CT),
\]
where $c,C>0$ are fixed. Moreover, for a fixed $c_0>0$,
\[
\min\{\dim_E(\mathcal G,d_\cdot,1/T),T\}
\le C\left\{\gamma_T^{\mathcal K}(c_0/T^2)
 +k\log(1+CT^3/k)\right\}.
\]
Here $\gamma_T^{\mathcal K}(\lambda)$ is the information gain of
$\mathcal K$ at regularization $\lambda$.
\end{lemma}
\begin{proof}
Take a sup-norm $c/T$-net of $\mathcal F$ and a Euclidean
$c/T$-net of the unit ball for $v$.
For each pair of net points, take the pointwise supremum of
$p_{v'}(r-f'(a)\mid a)$ over the corresponding net cell.
This function dominates every reward density in that cell.
The densities $p_v$ have uniformly bounded variation in $r$.
A shift of size $c/T$ therefore adds at most $C/T$ to the
integral of the supremum. Varying $v$ adds at most $C/T$.
Choose $c$ small enough that the combined excess is at most $1/T$.
There are at most $N(\mathcal F,c/T,\|\cdot\|_\infty)(CT)^k$
envelopes. This proves the first bound.

Write $\theta_v(a)=\langle v,\psi(a)\rangle$.
The proof of Lemma~\ref{lem:learned_cdf_example} shows, with fixed
positive constants $c_1,C_1$, that
\[
c_1\sqrt{(f(a)-f'(a))^2+(\theta_v(a)-\theta_{v'}(a))^2}
\le d_a(G_{f,v},G_{f',v'})
\le C_1\sqrt{(f(a)-f'(a))^2+(\theta_v(a)-\theta_{v'}(a))^2}.
\]
Use the two feature vectors
$(\mathcal K(\cdot,a),0)$ and $(0,\psi(a))$ in
$\mathcal H_{\mathcal K}\oplus\mathbb R^k$.
Their squared prediction errors add to the expression above.
For the difference of two parameters, the squared norm is at most
$4(B^2+1)$. Put $\lambda=c_0/T^2$ and define the block operator
\[
V_{i-1}=\operatorname{diag}\left(
\lambda I+\sum_{j<i}\phi(a_j)\otimes\phi(a_j),
\lambda I+\sum_{j<i}\psi(a_j)\psi(a_j)^\top\right),
\qquad \phi(a)=\mathcal K(\cdot,a).
\]
For an independent sequence at accuracy $\varepsilon'\ge1/T$,
its witness satisfies
\[
\|(f-f',v-v')\|_{V_{i-1}}^2
\le (4c_0(B^2+1)+c_1^{-2})(\varepsilon')^2.
\]
Write $g=f-f'$ and $z=v-v'$, and let
$V_{\phi,i-1}$ and $V_{\psi,i-1}$ denote the two diagonal blocks of
$V_{i-1}$. Define
\[
\ell_{\phi,i}
=
\|\phi(a_i)\|_{V_{\phi,i-1}^{-1}}^2,
\qquad
\ell_{\psi,i}
=
\|\psi(a_i)\|_{V_{\psi,i-1}^{-1}}^2.
\]
Applying Cauchy--Schwarz separately to the two prediction coordinates,
\begin{align*}
g(a_i)^2+\langle z,\psi(a_i)\rangle^2
&\le
\|g\|_{V_{\phi,i-1}}^2\ell_{\phi,i}
+\|z\|_{V_{\psi,i-1}}^2\ell_{\psi,i}
\\[-0.2em]
&\le
\|(g,z)\|_{V_{i-1}}^2
(\ell_{\phi,i}+\ell_{\psi,i}).
\end{align*}
On the other hand, the upper metric-comparison bound and
$d_{a_i}(G_{f,v},G_{f',v'})>\varepsilon'$ imply
$g(a_i)^2+\langle z,\psi(a_i)\rangle^2
>
\frac{(\varepsilon')^2}{C_1^2}.$

Combining the preceding inequalities gives

$\ell_{\phi,i}+\ell_{\psi,i}
>
\frac{1}{
C_1^2\bigl(4c_0(B^2+1)+c_1^{-2}\bigr)}.$

The corresponding increase in the sum of the two log determinants is
\begin{align*}
\log(1+\ell_{\phi,i})+\log(1+\ell_{\psi,i})
&\ge
\log(1+\ell_{\phi,i}+\ell_{\psi,i}),
\end{align*}
and is therefore bounded below by a fixed positive constant.
Over $n\le T$ independent arms, their total increase is at most
\[
2\gamma_T^{\mathcal K}(\lambda)
+k\log\left(1+\frac{T}{k\lambda}\right).
\]
For the finite block, use its rank at most $k$ and trace at most $T$.
For the RKHS block, use the Gram determinant identity.
This proves the bound for an independent sequence of length at most $T$.
If the dimension exceeds $T$, apply the argument to its first $T$ arms.
Substitute $\lambda=c_0/T^2$ to conclude.
\end{proof}

\begin{remark}[What the kernel bounds imply]
\label{rem:learned_cdf_kernel_rates}
The theorem now gives an explicit learned-CDF bound after
substituting the two estimates above.
For a kernel confidence band, this is
\begin{align*}
R(T)\le{}& C\left(1+\frac{p_{\max}}{p_{\min}}\right)
\sqrt{T\beta_T\gamma_T^{\mathcal K}(\lambda)}
+\frac{C}{p_{\min}}(1+D)\\
&+\frac{C}{p_{\min}}\sqrt{TD\left[
\log N(\mathcal F,c/T,\|\cdot\|_\infty)
+k\log(CT)+\log(2/\delta_0)\right]}.
\end{align*}
where $D$ is bounded by Lemma~\ref{lem:learned_cdf_kernel_example}.
The bound on $D$ uses information gain at $c_0/T^2$, not at the
constant regularization used for the known-CDF kernel algorithm.
Hence this argument alone does not preserve that algorithm's
Mat\'ern regret exponent. A sharper learned-CDF analysis would
be needed to make that claim.
\end{remark}

\begin{remark}[When learning preserves the rate]
The mean-width term has the same complexity dependence as in
known-CDF learning, up to the density constants.
The total rate is unchanged only when the added term fits within
that mean-width bound. For fixed $p_{\min},p_{\max}$, it suffices that
$1+D+\sqrt{D\beta^{\mathcal H}T}$ has at most that order.
The explicit linear family satisfies this condition when $k=O(d)$.
A finite-parameter noise class does not by itself establish this
condition for every kernel mean class.
\end{remark}

\begin{remark}[Scope of the learned-CDF guarantee]
The common support, density bounds, finite upper-density cover,
and distributional Eluder dimension are substantive assumptions.
The density lower bound converts a CDF error into a horizontal shift,
which permits discontinuous probability weights.
The theorem does not promise sublinear regret for every density class.
For fixed density bounds, sublinearity follows when the mean-width
sum is $o(T)$, $D=o(T)$, and $D\beta^{\mathcal H}=o(T)$.
\end{remark}

\section{Experiments}
\label{app:experiments}

Our experiments study the effect of the risk objective and the cost of
learning under that objective.
Part~I assumes known noise CDFs. It compares risk criteria under a common
practical exploration rule and evaluates \texttt{KCB} and \texttt{CPT-BPE} with their
analyzed confidence constructions, together with CPT-aware baselines,
on bounded kernel and Gaussian linear instances.
Part~II studies the finite-sample cost of learning unknown noise laws
using Algorithm~\ref{alg:npcb-learned}, with a known-law specialization
and a plug-in MLE baseline.

The experiments use distinct statistical models.
The practical comparison in Part~I uses a kernel-ridge mean estimator;
the additional kernel and linear instances evaluate the analyzed
known-CDF algorithms.
Part~II uses a finite realizable mean--noise class so that confidence-set
membership and joint optimistic optimization can be computed directly.

All experiments were run on a server with Ubuntu 24.04.3 LTS, an AMD EPYC 9124
16-core processor (32 hardware threads), and 125\,GiB of RAM.  The software
environment used Python 3.12.14 with NumPy 2.5.3, SciPy 1.18.1, and Matplotlib
3.11.2. The reported experiments use only CPU computation.

\subsection{Part I: Known-CDF Risk Criteria and Algorithms}
\label{app:exp-part1}

Part~I asks how risk criteria change the preferred arm and how CPT-aware
algorithms learn that arm. We use four constructions
(Table~\ref{tab:part1_setups}). The six-route construction compares Mean,
CVaR, mean--variance (MV), and CPT on common reward distributions.
The bounded kernel and Gaussian linear constructions evaluate the analyzed
algorithms and CPT-aware comparators. All regret uses the population CPT
objective with identity utility and the Tversky--Kahneman weight
$\mathbb w(p)=p^\gamma/(p^\gamma+(1-p)^\gamma)^{1/\gamma}$,
with reference parameter $\gamma=0.50$.

\begin{table}[t]
\centering
\caption{Part~I instances. A route groups nearby arms for reporting;
it is not a separate selectable object. Noise laws are supplied to the
known-CDF methods. W-UCB estimates each arm's reward distribution from samples.}
\label{tab:part1_setups}
\small
\setlength{\tabcolsep}{5pt}
\begin{tabular}{p{0.22\textwidth}p{0.16\textwidth}p{0.24\textwidth}p{0.28\textwidth}}
\toprule
Instance & Arms & Noise & Compared methods\\
\midrule
Six-route configurations A--E & $100$ in $[0,1]$ & Gaussian background; two-point tails & Mean, CVaR, MV, \texttt{KCB} ($2\sigma$)\\
Bounded 1D kernel & $50$ in $[0,1]$ & Centered two-point laws & \texttt{KCB}, \texttt{CPT-BPE}\\
Bounded 2D kernel & $7\times7$ in $[0,1]^2$ & Centered two-point laws & \texttt{KCB}, \texttt{CPT-BPE}, W-UCB\\

Gaussian linear
& $50$ arms with varying norms in $\mathbb R^3$
& Gaussian in-parameter noise
& Linear \texttt{KCB}, WOFUL, OFUL\\
\bottomrule
\end{tabular}
\end{table}

\paragraph{Six-route environment.}
The arms are $100$ equally spaced points in $[0,1]$.
An arm's route is the region of its nearest center among
$c=(0.10,0.25,0.40,0.55,0.70,0.85)$; R1--R6 denote these regions.
The mean is the sum of six Gaussian bumps,
$f^\star(a)=\sum_{j=1}^6 h_j\exp(-(a-c_j)^2/(2(0.03)^2))$.
Configs~A--D use $h=(10,8,10.6,6,4,2)$;
Config~E changes $h_3$ to $9.5$.
Arms in $[0.35,0.45]$ receive the rare-event noise law; all other arms have
$N(0,0.05^2)$ noise. The rare-event interval is a subset of the R3 reporting
region. For a downside event, noise is $-L$ with probability $p$ and
$pL/(1-p)$ otherwise. For an upside event, it is $L$ with probability $p$
and $-pL/(1-p)$ otherwise. Both laws are centered.
A--D use downside pairs $(p,L)=(0.001,15),(0.003,15),(0.018,5),(0.030,5)$;
E uses the upside pair $(0.010,100)$.

\paragraph{Criteria, practical decisions, and evaluation.}
Lower-tail CVaR is
$\operatorname{CVaR}_\alpha(X)=\alpha^{-1}\int_0^\alpha F_X^{-1}(u)\,du$,
with $F_X^{-1}(u)=\inf\{x:F_X(x)\ge u\}$; this definition handles atoms.
MV is $\operatorname{MV}_c(X)=\mathbb E[X]-c\operatorname{Var}(X)$.
The default parameters are $\alpha=0.10$, $c=1$, and $\gamma=0.50$.
Population scores use the true mean and noise laws.
Gaussian CVaR and variance are analytic, two-point scores are exact, and
Gaussian CPT uses $32{,}768$ deterministic midpoint quantiles.

For the practical comparison, each method uses kernel-ridge estimation with
squared-exponential lengthscale $0.05$, regularization $0.01$, and
$U_t(a)=\mu_{t-1}(a)+2\sigma_{t-1}(a)$.
It selects the arm maximizing its criterion for $U_t(a)+\epsilon_a$.
Each trial starts with six pulls at the grid arms nearest the route centers.
We run $60$ paired trials of length $T=200$.
The fixed multiplier $2$ is a practical exploration choice, rather than the
confidence multiplier of Algorithm~\ref{alg:kernel_cpt}; its recorded
all-arm mean-confidence coverage is zero.
We denote this implementation by \texttt{KCB} ($2\sigma$).

For selected arms $a_t$, cumulative CPT pseudo-regret is
$R_T=\sum_{t=1}^T[g(a^\star)-g(a_t)]$, where $g$ is the population
reference CPT value and $a^\star$ maximizes it on the actual arm grid.
A trial's final route is its modal route over the last $20\%$ of rounds.
Native success measures agreement with the learner's own population-optimal
route; CPT success measures agreement with the reference CPT-optimal route.
These are route-level diagnostics, not exact-arm identification guarantees.
When a native optimum has a positive CPT gap, a learner can accumulate linear
CPT regret despite learning its own objective successfully. Thus these
cross-objective curves measure the cost of decisions under the specified CPT
preference; the CPT-aware comparisons below assess learning on the CPT task.

\begin{table}[t!]
\centering
\caption{Population-optimal routes on the $100$-arm grid in Part~I.
CVaR uses $\alpha=0.10$, and CPT uses $\gamma=0.50$.
The MV column with $c=2$ supports the sensitivity discussion.
These are population optima, not learned final routes.}
\label{tab:population_objectives}
\small
\setlength{\tabcolsep}{5pt}
\begin{tabular}{cccccc}
\toprule
\textbf{Cfg}
& \textbf{Mean}
& \textbf{CVaR(.1)}
& \textbf{MV(1)}
& \textbf{MV(2)}
& \cellcolor{bestblue}\textbf{CPT(.5)}\\
\midrule
A & R3 & R3 & R3 & R3 & \cellcolor{bestblue}R1\\
B & R3 & R3 & R1 & R1 & \cellcolor{bestblue}R1\\
C & R3 & R1 & R3 & R1 & \cellcolor{bestblue}R1\\
D & R3 & R1 & R1 & R1 & \cellcolor{bestblue}R1\\
E & R1 & R1 & R1 & R1 & \cellcolor{bestblue}R3\\
\bottomrule
\end{tabular}
\end{table}

\begin{table}[t]
\centering
\caption{Practical six-route comparison over $60$ paired trials at $T=200$.
Regret entries are cumulative reference-CPT regret, mean $\pm$ one standard
error. Success entries are native/CPT route-success percentages.
All methods use the common $2\sigma$ exploration rule.
Native optima are given in Table~\ref{tab:population_objectives}.}
\label{tab:main_results}
\small
\setlength{\tabcolsep}{6pt}
\begin{tabular}{clccc>{\columncolor{bestblue}}c}
\toprule
Cfg & Metric & Mean-UCB & CVaR-UCB & MV-UCB & \texttt{KCB} ($2\sigma$)\\
\midrule
A & Regret & $63.5\pm0.38$ & $63.2\pm0.69$ & $63.2\pm0.69$ & $24.2\pm0.03$\\
  & Success (native/CPT) & $97/3$ & $95/5$ & $95/5$ & $100/100$\\
\midrule
B & Regret & $151.6\pm3.35$ & $143.2\pm5.08$ & $25.7\pm0.03$ & $22.0\pm0.03$\\
  & Success (native/CPT) & $95/5$ & $88/12$ & $100/100$ & $100/100$\\
\midrule
C & Regret & $99.5\pm4.39$ & $25.2\pm0.08$ & $42.0\pm4.54$ & $21.9\pm0.06$\\
  & Success (native/CPT) & $82/18$ & $100/100$ & $18/82$ & $100/100$\\
\midrule
D & Regret & $142.0\pm4.87$ & $21.9\pm0.02$ & $25.8\pm0.10$ & $21.9\pm0.02$\\
  & Success (native/CPT) & $90/10$ & $100/100$ & $100/100$ & $100/100$\\
\midrule
E & Regret & $1362.2\pm18.32$ & $1372.0\pm11.00$ & $1387.8\pm0.66$ & $58.6\pm1.96$\\
  & Success (native/CPT) & $100/0$ & $100/0$ & $100/0$ & $100/100$\\
\bottomrule
\end{tabular}
\end{table}

\begin{figure}[t]
\centering
\includegraphics[width=\textwidth,height=0.88\textheight,keepaspectratio]{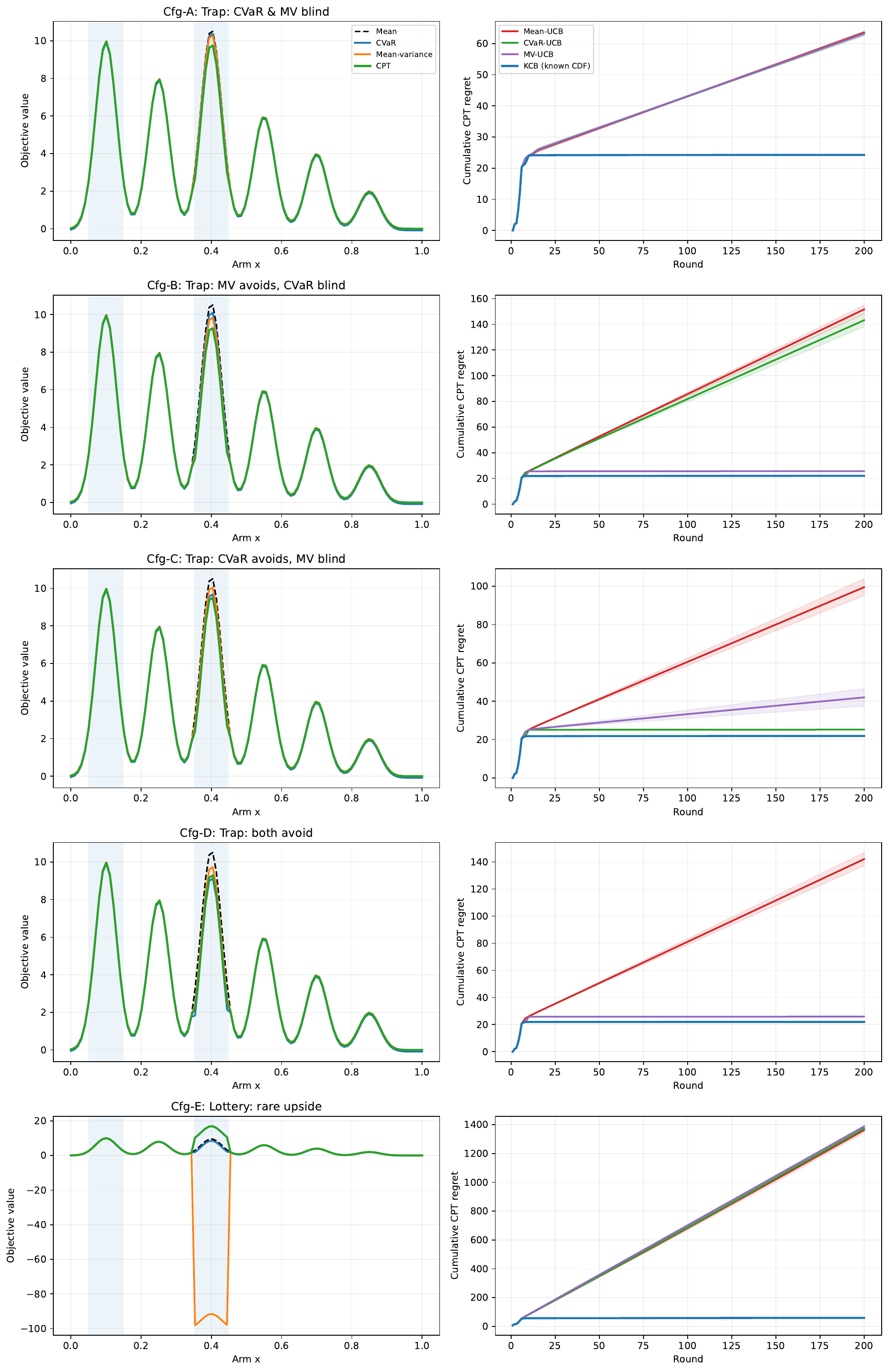}
\caption{Six-route objective landscapes and practical decisions.
Each row is a configuration A--E. Left: population Mean, CVaR, MV, and CPT
scores, computed from the true reward laws independently of exploration.
Right: cumulative reference-CPT regret over $60$ paired trials, with one
standard-error shading. All right-panel methods use the common $2\sigma$
rule; the \texttt{KCB} legend here denotes \texttt{KCB} ($2\sigma$).
These curves illustrate objective disagreement and practical learning
behavior, rather than validate the analyzed \texttt{KCB} confidence construction.}
\label{fig:risk_landscapes}
\end{figure}

The population comparisons explain several different preferences
(Table~\ref{tab:population_objectives}). In A, Mean, CVaR, and MV prefer R3
while CPT prefers R1; B and C separate CVaR and MV in opposite directions.
In E, R3 has lower mean and larger variance than R1 but a rare large upside:
Mean, lower-tail CVaR, and MV prefer R1, whereas CPT prefers R3.
The practical CPT learner reaches its reference-optimal route in every
default-parameter trial. Native success prevents other learners' CPT regret
from being mistaken for unsuccessful native-objective learning; for example,
MV-UCB in C reaches R1 in $82\%$ of trials although its native optimum is R3.

\paragraph{Parameter sensitivity.}
\label{subsec:sensitivity}
We also test $\alpha\in\{0.05,0.10,0.20\}$,
$c\in\{0.5,1,2\}$, and $\gamma\in\{0.4,0.5,0.6\}$.
Changing a parameter changes the corresponding population objective.
In B, CVaR prefers R1 at $\alpha=0.05$ and R3 at $0.10$ and $0.20$;
in C it prefers R1 at $0.05$ and $0.10$, and R3 at $0.20$.
In A, MV still prefers R3 at $c=2$ (best scores $10.0539$ versus $9.9894$
on R1), although its practical learner finishes on R1 in all trials.
CPT in A prefers R1 at $\gamma=0.4$ and $0.5$, but R3 at $0.6$;
the latter runs have $80\%$ native success and $20\%$ reference-CPT success.
In E, all tested CVaR and MV parameters prefer R1, while all tested CPT
parameters prefer R3; the corresponding practical learners reach those
routes in every trial. These findings concern specified preferences and
finite-horizon decisions, rather than universal dominance or a human-subject
validation of the chosen CPT parameters.

\paragraph{Kernel instances and theoretical confidence rules.}
The bounded 1D instance has $50$ equally spaced arms and
$k(a,b)=\exp(-(a-b)^2/(2(0.12)^2))$.
Its mean is $f^\star(a)=0.54k(a,0.25)+0.60k(a,0.72)$.
Arms in $[0.62,0.82]$ have noise $-0.30$ with probability $0.05$ and
$0.05\cdot0.30/0.95$ otherwise; other arms have symmetric noise $\pm0.02$.
The explicit kernel-section representation certifies $B=0.8074$,
Hoeffding's lemma gives $\eta=0.1579$, and all rewards lie in
$[-M,M]$ with $M=0.6154$.
The mean/CPT optima are $0.714/0.245$ on the grid.

The bounded 2D instance uses a $7\times7$ grid,
$k_0(a,b)=\exp(-\|a-b\|^2/(2(0.16)^2))$, and
$\mathcal K=0.20+0.80k_0$.
With $c_s=(0.25,0.25)$ and $c_r=(0.75,0.75)$, its mean is
$f^\star(a)=0.10+0.54k_0(a,c_s)+0.60k_0(a,c_r)$.
Arms within distance $0.17$ of $c_r$ use the risky law above; the others
have noise $\pm0.02$. The direct-sum kernel representation gives
\begin{align*}
B^2=\frac{0.10^2}{0.20}
+\frac{0.54^2+0.60^2+2(0.54)(0.60)k_0(c_s,c_r)}{0.80}.
\end{align*}
Thus $B=0.9298$ and $\eta=0.1579$.
All reward endpoints lie in $[0,M]$, with $M$ the largest endpoint on the grid.
The mean/CPT optima are $(2/3,2/3)/(1/3,1/3)$.

On both instances, \texttt{KCB} implements Algorithm~\ref{alg:kernel_cpt} with
$\lambda=1+2/T$, $\delta=0.10$, and its theoretical confidence multiplier:
\begin{align*}
\sqrt{\beta_t}
&=B+\eta\sqrt{2\bigl(\bar\gamma_{t-1}(\lambda)+1+\log(1/\delta)\bigr)},\\
\bar\gamma_n(\lambda)
&=\frac{r_n}{2}\log\!\left(1+\frac{n}{r_n\lambda}\right),
\quad r_n=\min\{n,r\},\quad n\ge1.
\end{align*}
Here $\bar\gamma_0=0$ and $r$ is the finite kernel matrix rank bound.
The trace and rank bounds imply $\gamma_n(\lambda)\le\bar\gamma_n(\lambda)$
for every design, since $\mathcal K(a,a)\le1$.
\texttt{KCB} selects the arm maximizing the supplied-law CPT value at
$\mu_{t-1}+\sqrt{\beta_t}\sigma_{t-1}$, with no fitted confidence scale.
\texttt{CPT-BPE} implements Algorithm~\ref{alg:cpt-bpe} with
$n_0=1$, $n_i=\lceil\sqrt{Tn_{i-1}}\rceil$, and
$c_T=B+(\eta/\sqrt\lambda)\sqrt{2\log(2m|\mathcal A|/\delta)}$,
where $m$ is the prescribed number of phases.
Each phase's design uses posterior variance and surviving arms, independently
of current-phase rewards. Estimates use only that phase's observations;
CPT upper/lower scores determine elimination at its end.
Both constructions have centered bounded noise and valid norm/noise
certificates, satisfying the assumptions of the compared kernel procedures.

\paragraph{CPT-aware comparators.}
In 2D we additionally run empirical-CDF W-UCB
\cite{DBLP:journals/corr/GopalanAFM16}.
It pulls every arm once, then maximizes its empirical CPT value plus
$L_wM(3\log t/(2n_a))^{\alpha_w/2}$, using the valid conservative weight
constants $\alpha_w=0.50$ and $L_w=5$.
The empirical CPT calculation handles atoms and ties directly.
W-UCB uses neither the kernel nor the supplied noise CDFs, and learns each
arm's distribution separately; \texttt{KCB} and \texttt{CPT-BPE} form the matched known-CDF
kernel comparison.

The linear comparison uses $50$ arms in $\mathbb R^3$ with
heterogeneous norms and $\theta_\star=(2,0,0)$.
Rewards follow $r_t=x_t^\top(\theta_\star+N_t)$, where
$N_t\sim\mathcal N(0,I_3)$ independently across rounds.
Thus the noise standard deviation at arm $x$ is $\|x\|_2$.
For linear KCB, we use $B=\|\theta_\star\|_2=2$ and
$\eta=\max_a\|x_a\|_2=1$.

The CPT-optimal arm is $x_{\rm CPT}=(0.300,0,0)$.
The mean-optimal arm is
$x_{\rm mean}=(0.305,\sqrt{1-0.305^2},0)$.
Their means are $0.600$ and $0.610$, respectively,
but their CPT values at $\gamma=0.50$ are approximately
$0.4628$ and $0.4355$.
The remaining arms comprise four coordinate-direction controls
and $44$ deterministically generated arms with norms in
$[0.35,0.95]$ and means below $0.58$.
We compare linear \texttt{KCB}, WOFUL, and OFUL on this same instance.

The arm norms vary, so the projected Gaussian noise has
arm-dependent variance.
Consequently, CPT scores need not preserve mean ordering,
and WOFUL need not select the same arms as OFUL.
We include OFUL as a risk-neutral comparison and evaluate
all three methods using the same CPT regret.
This Gaussian comparison uses the sub-Gaussian \texttt{KCB} setting and
finite Gaussian CPT integrals; the \texttt{CPT-BPE} comparisons use bounded rewards.

\begin{figure}[t]
\centering
\begin{minipage}[t]{0.325\textwidth}
    \centering
    \includegraphics[width=\linewidth]{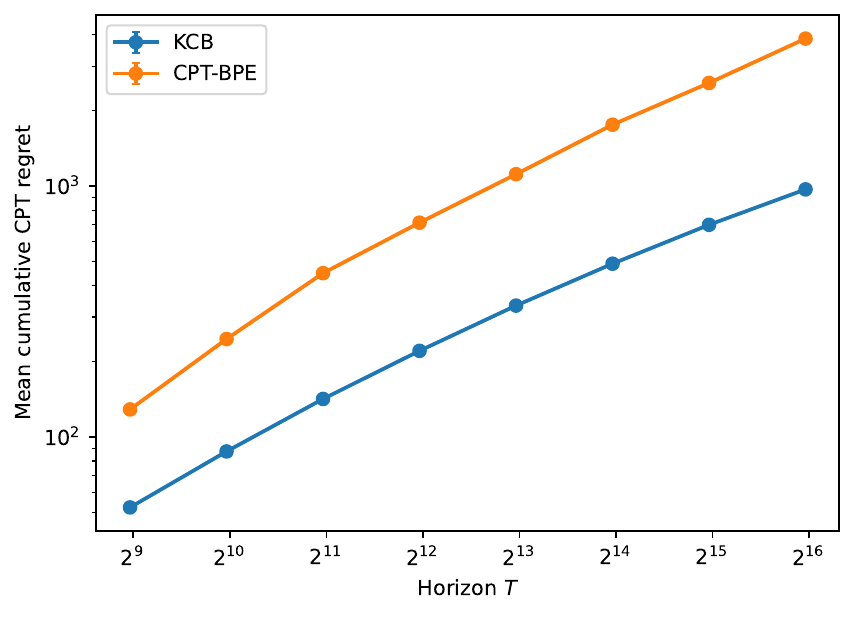}

    \small (a) Bounded 1D kernel
\end{minipage}
\hfill
\begin{minipage}[t]{0.325\textwidth}
    \centering
    \includegraphics[width=\linewidth]{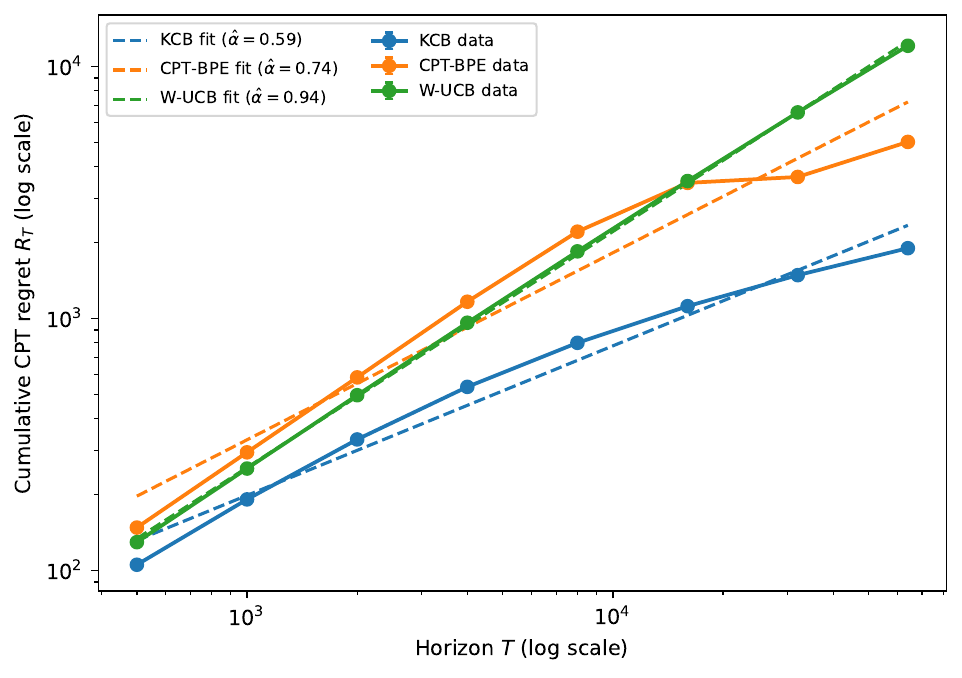}

    \small (b) Bounded 2D kernel
\end{minipage}
\hfill
\begin{minipage}[t]{0.325\textwidth}
    \centering
    \includegraphics[width=\linewidth]{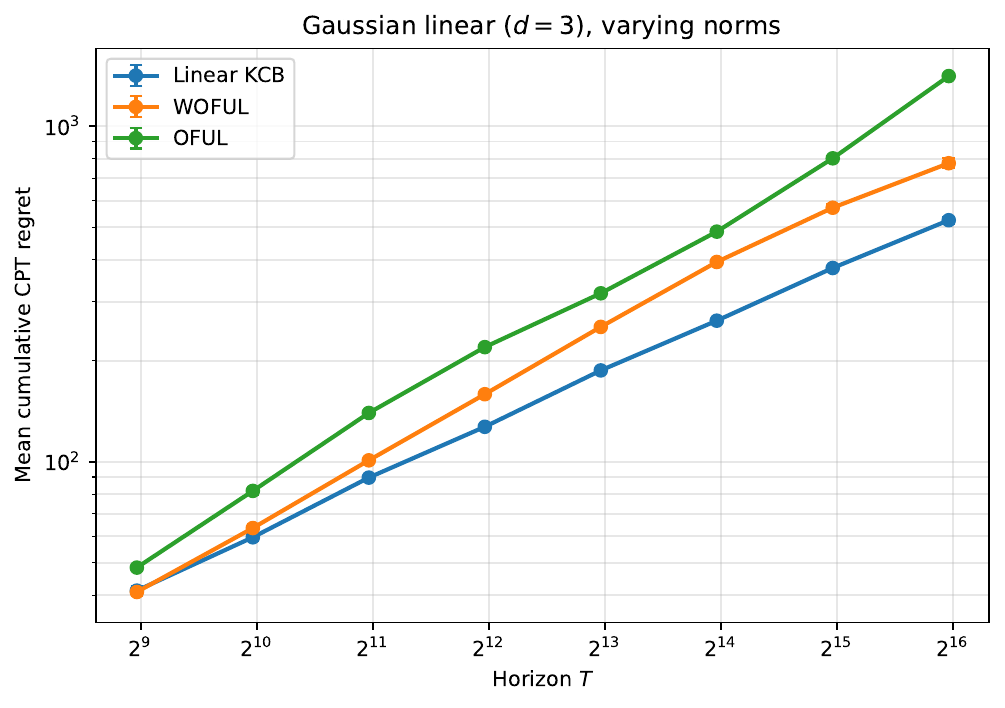}

    \small (c) Linear instance, $d=3$
\end{minipage}
\caption{
Cumulative CPT regret across eight horizons from $T=500$ to $64000$,
with logarithmic axes.
(a) \texttt{KCB} and \texttt{CPT-BPE} on the bounded 1D kernel instance.
(b) \texttt{KCB}, \texttt{CPT-BPE}, and empirical-CDF W-UCB on the bounded 2D kernel instance.
(c) Linear \texttt{KCB}, WOFUL, and OFUL on the Gaussian
linear instance with varying arm norms.
\texttt{KCB} and \texttt{CPT-BPE} use their respective analyzed confidence constructions.
Each horizon is run independently with its horizon-dependent parameters.
Dashed lines in (b) are descriptive log--log fits, not theoretical rates.
}
\label{fig:part1_algorithm_scaling}
\end{figure}

\begin{figure}[t]
\centering
\includegraphics[width=.8\textwidth]{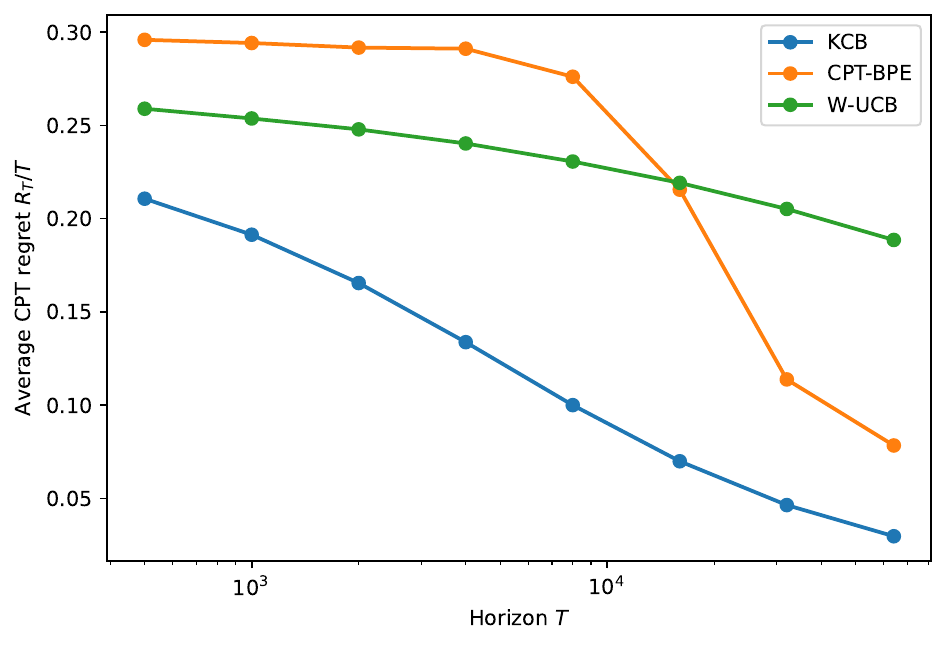}
\caption{
CPT regret per round, $R_T/T$, on the bounded 2D kernel instance.
The horizontal axis is logarithmic.
Average regret decreases across the tested horizons for all three methods,
with a pronounced reduction for \texttt{CPT-BPE} at larger horizons.
At $T=64000$, mean regret per round is $0.030$ for \texttt{KCB},
$0.078$ for \texttt{CPT-BPE}, and $0.189$ for empirical-CDF W-UCB.
}
\label{fig:kernel2d_average_regret}
\end{figure}

\begin{table}[t]
\centering
\caption{Cumulative CPT regret, mean $\pm$ one standard error
over $30$ paired trials at each horizon.
\texttt{KCB} and \texttt{CPT-BPE} use their analyzed confidence constructions,
rather than the practical $2\sigma$ rule in
Table~\ref{tab:main_results}.
W-UCB estimates each arm's CDF from observations.
Values are rounded to two decimal places.}
\label{tab:analyzed_cpt_results}
\footnotesize
\setlength{\tabcolsep}{3pt}
\renewcommand{\arraystretch}{1.12}
\begin{tabular}{
r
>{\columncolor{bestblue}}c
>{\columncolor{goodgreen}}c
>{\columncolor{bestblue}}c
>{\columncolor{goodgreen}}c
c
>{\columncolor{bestblue}}c
>{\columncolor{goodgreen}}c
}
\toprule
& \multicolumn{2}{c}{Bounded 1D kernel}
& \multicolumn{3}{c}{Bounded 2D kernel}
& \multicolumn{2}{c}{Gaussian linear ($d=3$)}\\
\cmidrule(lr){2-3}
\cmidrule(lr){4-6}
\cmidrule(lr){7-8}
$T$
& \texttt{KCB} & \texttt{CPT-BPE}
& \texttt{KCB} & \texttt{CPT-BPE} & W-UCB
& Linear \texttt{KCB} & WOFUL\\
\midrule
500
& $52.30\pm0.09$ & $128.52\pm0.00$
& $105.34\pm0.13$ & $147.95\pm0.00$ & $129.47\pm0.16$
& $57.83\pm1.87$ & $51.51\pm1.23$\\
1000
& $87.41\pm0.12$ & $245.23\pm0.00$
& $191.38\pm0.12$ & $294.18\pm0.00$ & $253.71\pm0.29$
& $83.61\pm2.51$ & $79.29\pm2.38$\\
2000
& $141.53\pm0.13$ & $448.93\pm0.55$
& $330.94\pm0.21$ & $583.45\pm0.00$ & $495.74\pm0.24$
& $118.21\pm2.60$ & $109.92\pm2.43$\\
4000
& $220.06\pm0.19$ & $713.56\pm0.00$
& $535.00\pm0.32$ & $1164.53\pm0.00$ & $961.21\pm0.25$
& $156.28\pm2.45$ & $143.74\pm2.84$\\
8000
& $333.33\pm0.34$ & $1113.10\pm1.85$
& $799.93\pm0.41$ & $2208.74\pm4.81$ & $1845.01\pm0.68$
& $215.71\pm4.40$ & $198.87\pm4.02$\\
16000
& $489.65\pm0.56$ & $1753.34\pm5.33$
& $1118.13\pm0.49$ & $3448.40\pm0.00$ & $3507.00\pm0.98$
& $289.04\pm6.45$ & $259.77\pm5.84$\\
32000
& $699.68\pm0.89$ & $2571.92\pm0.00$
& $1484.38\pm0.67$ & $3640.87\pm0.00$ & $6565.94\pm1.32$
& $356.77\pm7.55$ & $307.62\pm7.03$\\
64000
& $968.49\pm1.35$ & $3864.18\pm4.73$
& $1899.19\pm1.19$ & $5014.49\pm0.00$ & $12068.36\pm2.42$
& $437.36\pm10.07$ & $365.21\pm8.61$\\
\bottomrule
\end{tabular}
\end{table}

\paragraph{Algorithmic results.}
Table~\ref{tab:analyzed_cpt_results} reports the numerical results
corresponding to Figure~\ref{fig:part1_algorithm_scaling}.
Figure~\ref{fig:kernel2d_average_regret} additionally shows
the decrease in regret per round on the bounded 2D instance.
We run $30$ paired trials at each
$T\in\{500,1000,2000,4000,8000,16000,32000,64000\}$,
with independently configured horizon-dependent parameters.
\texttt{KCB} has lower mean regret than \texttt{CPT-BPE} on both bounded instances
at every tested horizon.
In 1D, regret per round decreases from $0.105$ to $0.015$
for \texttt{KCB} and from $0.257$ to $0.060$ for \texttt{CPT-BPE}.
In 2D, it decreases from $0.211$ to $0.030$ for \texttt{KCB},
from $0.296$ to $0.078$ for \texttt{CPT-BPE}, and from $0.259$ to $0.189$
for W-UCB.
\texttt{CPT-BPE} exhibits a pronounced slowdown at larger horizons
and overtakes W-UCB at $T=16000$.
The descriptive log--log slopes over the combined 2D range
are $0.593$, $0.742$, and $0.937$, respectively;
these fits describe finite-horizon growth rather than establish
asymptotic theoretical exponents.

On the varying-norm linear instance, WOFUL has slightly lower
mean regret at $T=500$, while linear \texttt{KCB} has lower
mean regret at every larger tested horizon.
Both CPT-aware methods have lower mean CPT regret than OFUL
throughout the tested range.
Their descriptive log--log slopes are $0.527$ for linear
\texttt{KCB}, $0.621$ for WOFUL, and $0.672$ for OFUL.

Recorded \texttt{KCB} all-arm mean-confidence coverage and \texttt{CPT-BPE}
phase-end mean-confidence coverage are $1.000$ on the bounded runs.

On the varying-norm linear instance, the recorded mean-confidence
coverage is $1.000$ in every trial for all three methods.
These are observed confidence diagnostics for the tested runs.
Some \texttt{CPT-BPE} standard errors round to zero because trials can
produce identical sampling sequences when their phase-end
elimination decisions coincide.
\subsection{Part II: Learning the Noise CDF}
\label{app:exp-part2}

Part~II measures the additional learning cost when the regional noise laws
are unknown. We evaluate NPCB-Learned
(Algorithm~\ref{alg:npcb-learned}) in a finite realizable mean--noise class,
where confidence-set membership and optimistic maximization can be computed
directly. This experiment tests the finite-class algorithm rather than the
computational scalability or regret rates of the continuous kernel extension.

\paragraph{Environment and objectives.}
The selectable arms are $100$ equally spaced points $x_a\in[0,1]$.
An arm belongs to the route with nearest center among
$(0.10,0.25,0.40,0.55,0.70,0.85)$; routes group arms for reporting and are
not separate selectable objects. For region $r(a)$, the mean is
$f_\theta(a)=\theta_{r(a)}\phi_a$, where
$\phi_a=\exp(-(x_a-c_{r(a)})^2/(2(0.04)^2))$.
The true route heights are $(10,8.5,10.6,7,5,3.5)$ in A--D and
$(10,8.5,9.5,7,5,3.5)$ in E. Route~3 carries the rare-event noise;
the other routes use the no-tail law described below.
In A--D, rare losses make Route~1 CPT-optimal despite Route~3's higher mean.
In E, rare gains make Route~3 CPT-optimal despite its lower mean.
All methods are evaluated using cumulative population CPT regret
$R_T=\sum_{t=1}^T[g(a^*)-g(a_t)]$, with identity utility and the reference
Tversky--Kahneman weight parameter $\gamma=0.50$.

\paragraph{Finite mean and noise classes.}
Each candidate route height belongs to
$\Theta=\{0,0.1,\ldots,14\}$, giving $|\mathcal F|=141^6$.
The true route-height vector lies in this product grid.
Each regional noise law belongs to a family with
$p\in\{0.001,0.002,0.003,0.005,0.010,0.018,0.030,0.050\}$,
$L\in\{5,10,15,30,60,100\}$, and a no-tail law.
The tail sign is supplied: downside in A--D and upside in E.
Thus each region has $49$ candidate laws and $|\mathcal H|=49^6$;
the regional assignments, tail probabilities, and magnitudes are unknown.

The laws are continuous densities on the common support $[-s,s]$, with
$s=106$. Each combines a uniform component of mass $\kappa=10^{-6}$
with two truncated, normalized Gaussian components of standard deviation
$0.1$. For a downside law, the component centers are $pL/(1-p)$ and $-L$,
with respective mixture weights $(1-\kappa)(1-p)$ and $(1-\kappa)p$;
for an upside law, both centers change sign.
The no-tail law has one Gaussian component centered at zero and the same
uniform component. The untruncated mixtures are centered; the truncation
changes their means by less than floating-point resolution at these settings.
Sampling, likelihood evaluation, and CPT evaluation use the same normalized
compact-support law.
The density bounds are $p_{\min}=\kappa/(2s)\simeq4.72\times10^{-9}$
and the valid common upper bound $p_{\max}=4$.
Every candidate mean lies in $[0,14]$, so every candidate reward lies
in $[-106,120]\subseteq[-120,120]$.

\paragraph{Methods and implementation.}
We compare four methods using the same six-round initialization,
one pull at the arm nearest each route center.
\texttt{Mean-UCB} maximizes the optimistic mean over the joint likelihood
and mean confidence sets.
\texttt{NPCB (known law)} is the known-law specialization of the finite-class
CPT learner: the true regional noise laws are supplied, while the means
remain unknown.
\texttt{NPCB-Learned} learns both components and maximizes CPT value over
the joint feasible set.
\texttt{Plug-in MLE} selects an arm using CPT values at the current
regional joint maximum-likelihood estimates, without a confidence-set
optimism bonus. It uses the same candidate mean and noise families.
This baseline isolates point-estimate decisions from optimistic decisions
within the specified model family; it is not a nonparametric empirical-CDF
estimator.

For a global pair $(f,h)$, the log-likelihood factorizes as
$\ell_t(f,h)=\sum_{r=1}^6\ell_{t,r}(\theta_r,h_r)$.
The likelihood set is
$\mathcal M_t=\{(f,h):\ell_t(f,h)\ge\sup_{f',h'}\ell_t(f',h')-
\beta^{\mathcal H}\}$.
We use $\delta_0=0.05$ and the finite-class radius
$\beta^{\mathcal H}=\log(2|\mathcal F||\mathcal H|/\delta_0)+1=57.732$.
The finite shifted-density class supplies a valid upper cover with
$N_{\mathrm{opt}}(\mathcal Q,1/T)\le|\mathcal F||\mathcal H|$;
this need not be the smallest cover.
For the known-law specialization, the noise class is a singleton,
so the corresponding radius is
$\beta^{\mathcal H}
=\log(2|\mathcal F|/\delta_0)+1=34.381$.
The mean set $\mathcal C_t$ is the product of regional grid sets whose
empirical squared losses are within
$8s^2\log(6|\Theta|/\delta_0)$ of the corresponding regional least-squares
minimum.

For each arm, NPCB-Learned computes
$\widetilde{\mathcal V}_t(a)=
\sup\{\mathcal V_{h_{r(a)}}(f(a)):(f,h)\in\mathcal M_t,
 f\in\mathcal C_t\}$.
Regional factorization permits this optimization without enumerating all
$141^6 49^6$ global pairs: other regions are assigned their best
mean-feasible likelihood values when testing a candidate regional pair.
The CPT score uses the same mean--noise pair that passes both constraints.
The implementation reconstructs a full feasible global witness for the
selected arm each round.
Finite-set optimization is exact; CPT integrals are evaluated using
deterministic quadrature and interpolation tables, with integration and
mean-grid spacings $0.005$.

\paragraph{Trials and reporting.}
The horizons and true tail parameters are listed in
Table~\ref{tab:learned_results}.
We use $30$ paired trials with base seed $0$.
For each configuration and trial, a common table of potential noise
realizations for every arm and round is supplied to all methods.
A trial's final route is its modal route over the last $20\%$ of rounds.
Learned success is the fraction of NPCB-Learned trials whose final route
is CPT-optimal. This route statistic is distinct from exact-arm identification.
Table entries report mean regret and one standard error;
curve shading reports one standard deviation, showing variation across trials.

\begin{table}[t!]
\centering
\caption{Part~II: cumulative CPT regret, mean $\pm$ one standard error
over $30$ paired trials. Parentheses give the most frequent final route
across trials; each trial's final route is measured over its last $20\%$
of rounds. The final column reports CPT-optimal final-route success
for plug-in MLE/NPCB-Learned. A--D have rare losses; E has rare gains.}
\label{tab:learned_results}
\small
\setlength{\tabcolsep}{2pt}
\begin{tabular}{@{}ccccccccc@{}}
\toprule
Cfg & $p$ & $L$ & $T$
& \texttt{Mean-UCB}
& \shortstack{\texttt{NPCB}\\(known law)}
& \shortstack{\texttt{Plug-in}\\MLE}
& \texttt{NPCB-Learned}
& \shortstack{Success (\%)\\plug-in/learned}\\
\midrule
A & 0.001 & 15 & 20000
& \cellcolor{badred}$3572.2\pm0.6$ (R3)
& \cellcolor{goodgreen}$35.2\pm0.1$ (R1)
& \cellcolor{badred}$2692.1\pm248.3$ (R3)
& \cellcolor{bestblue}$901.7\pm50.7$ (R1)
& $20.0/100$\\
B & 0.003 & 15 & 10000
& \cellcolor{badred}$6610.2\pm0.6$ (R3)
& \cellcolor{goodgreen}$33.4\pm0.0$ (R1)
& \cellcolor{badred}$5354.7\pm442.8$ (R3)
& \cellcolor{bestblue}$681.3\pm34.2$ (R1)
& $16.7/100$\\
C & 0.018 & 5 & 5000
& \cellcolor{badred}$2323.1\pm0.8$ (R3)
& \cellcolor{goodgreen}$33.6\pm0.1$ (R1)
& \cellcolor{badred}$1846.2\pm133.0$ (R3)
& \cellcolor{bestblue}$821.5\pm29.7$ (R1)
& $13.3/100$\\
D & 0.030 & 5 & 5000
& \cellcolor{badred}$3296.7\pm0.8$ (R3)
& \cellcolor{goodgreen}$33.4\pm0.0$ (R1)
& \cellcolor{badred}$2701.0\pm195.4$ (R3)
& \cellcolor{bestblue}$652.4\pm14.4$ (R1)
& $13.3/100$\\
E & 0.010 & 100 & 30000
& \cellcolor{badred}$211787.4\pm214.7$ (R1)
& \cellcolor{goodgreen}$50.1\pm0.0$ (R3)
& \cellcolor{badred}$204225.3\pm0.9$ (R1)
& \cellcolor{bestblue}$98458.2\pm887.7$ (R3)
& $0.0/100$\\
\bottomrule
\end{tabular}
\end{table}

\begin{figure}[t!]
\centering
\includegraphics[width=\textwidth,height=0.82\textheight,keepaspectratio]{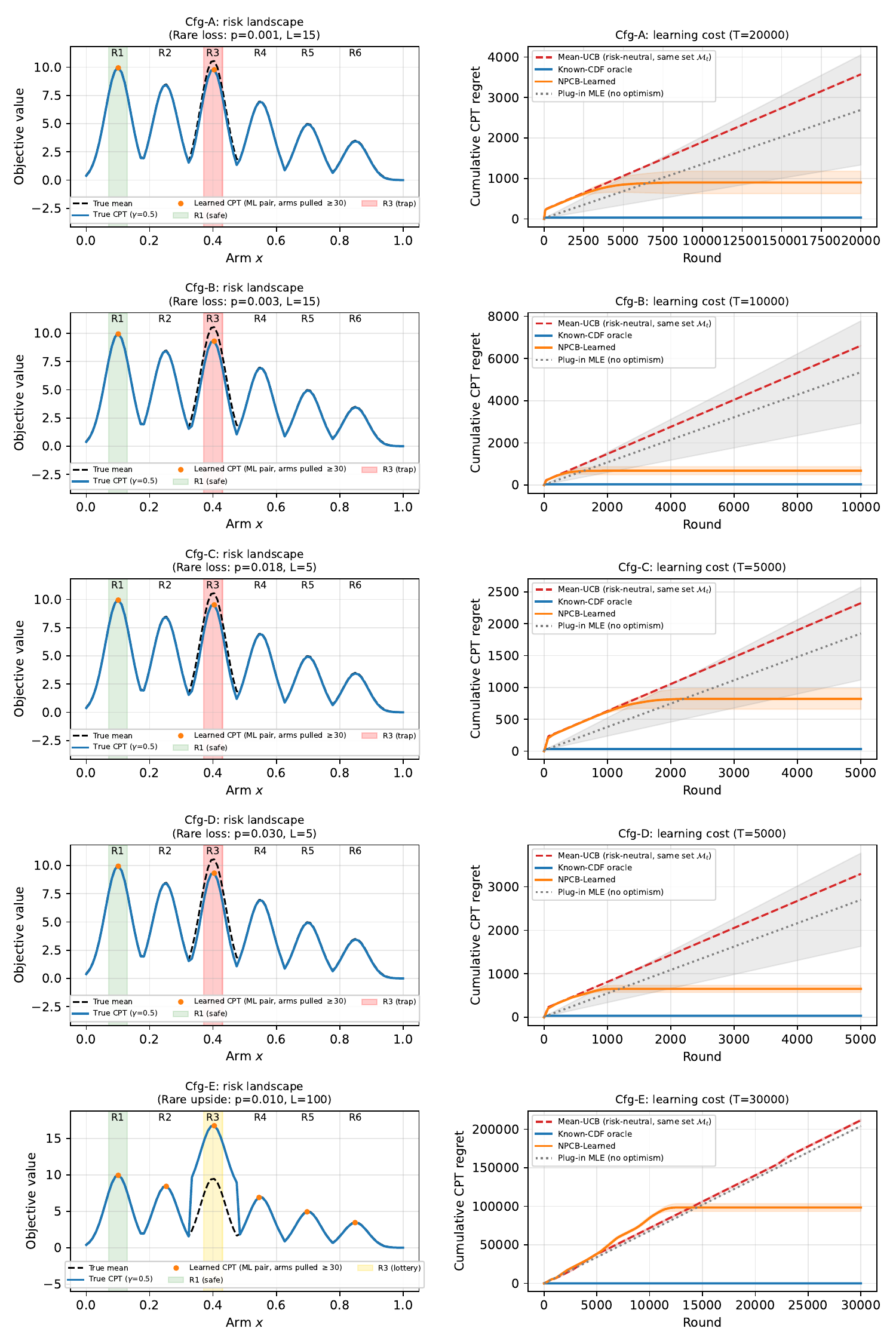}
\caption{\textbf{Part II: unknown-noise learning.}
Rows correspond to A--E.
Left: true mean and CPT landscapes, with fitted CPT values from the joint
maximum-likelihood model on arms pulled at least $30$ times in a
representative NPCB-Learned trial.
The representative trial is the successful trial nearest the median
final regret. These fitted values are diagnostics, not the optimistic
scores used for selection or a full-domain CDF accuracy assessment.
Right: mean cumulative CPT regret for Mean-UCB, the known-law specialization,
NPCB-Learned, and plug-in MLE; shading is one standard deviation across
$30$ trials.}
\label{fig:online_learnt_cdf}
\end{figure}

\paragraph{Learning cost and the plug-in comparison.}
The known-law specialization quickly selects the CPT-preferred route and
incurs only a small initialization and mean-learning cost.
NPCB-Learned incurs additional regret while resolving uncertainty about
the regional noise laws, then selects the CPT-preferred route in all
recorded trials. In A--D, its final mean regret ranges from $652$ to $902$,
compared with approximately $33$--$35$ when the true laws are supplied.
The especially rare downside event in A has probability $10^{-3}$;
observing such an event requires substantial sampling of its region.

E illustrates a larger finite-sample cost.
NPCB-Learned incurs mean regret $98458$, compared with $50.1$ for the
known-law specialization and $211787$ for Mean-UCB, while reaching the
CPT-optimal final route in all trials.
The route-convergence curves show that this outcome follows a prolonged
learning period rather than immediate identification.

NPCB-Learned has lower mean CPT regret than plug-in MLE in every
configuration (Table~\ref{tab:learned_results}).
Its CPT-optimal final-route success is $100\%$ throughout,
whereas plug-in MLE achieves $20.0\%$, $16.7\%$, $13.3\%$,
$13.3\%$, and $0\%$ in A--E, respectively.
Because both methods optimize CPT under the same unknown-law model,
this comparison distinguishes confidence-set optimism from
point-estimate decisions rather than disagreement between risk objectives.
This observation concerns this plug-in rule and these instances; it does
not establish superiority over all distribution-learning methods.
Mean-UCB optimizes a different objective, so its continuing CPT regret
primarily reflects the preference reversal rather than failed mean learning.

\begin{figure}[t!]
\centering

\includegraphics[width=0.32\textwidth]{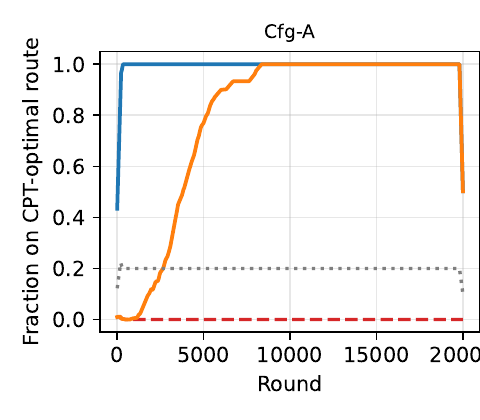}\hfill
\includegraphics[width=0.32\textwidth]{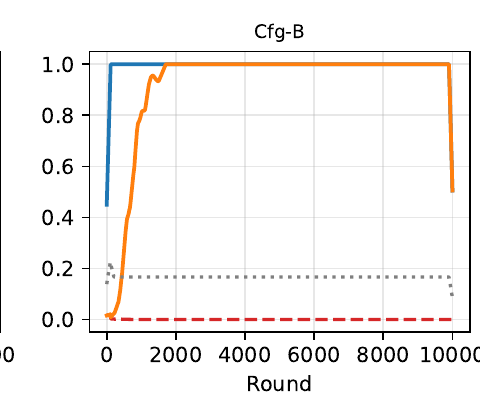}\hfill
\includegraphics[width=0.32\textwidth]{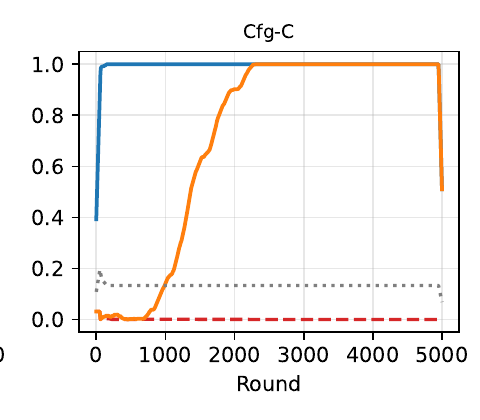}

\vspace{4pt}

\includegraphics[width=0.32\textwidth]{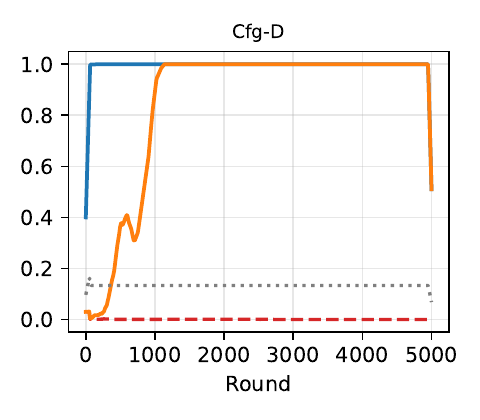}\hspace{0.03\textwidth}
\includegraphics[width=0.32\textwidth]{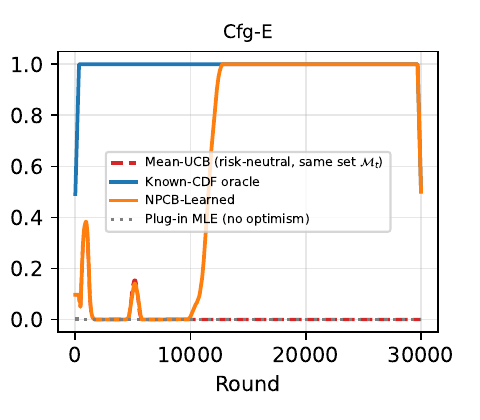}
\caption{\textbf{Selection of the CPT-optimal route in Part II.}
Each panel shows the smoothed fraction of the $30$ trials selecting the
CPT-optimal route at each round for the four methods.
Smoothing is used only for visualization; route-success and last-selection
statistics are calculated from the unsmoothed trajectories.}
\label{fig:part2_convergence}
\end{figure}
For NPCB-Learned, the median last round selecting a route other than the
CPT-optimal route is approximately $3865$, $788$, $1461$, $848$, and $11519$
in A--E, respectively; the corresponding maxima over trials are
$8225$, $1608$, $2255$, $1085$, and $12446$.
All are before the final $20\%$ of their respective horizons.
The recorded regret trajectories also plateau over that final interval.
Route selection alone would not imply zero regret within a region;
the plateau is a separate observation from the arm-level regret records.

\paragraph{Confidence and feasibility diagnostics.}
For the confidence-set methods, the recorded fraction of rounds in which
the generating model belongs to both confidence sets is $1.000$ in every
configuration and trial.
For NPCB-Learned, the reconstructed global witness passes the likelihood
and mean-confidence constraints at every selected arm; the recorded number
of feasibility violations is zero.
These are sample-path diagnostics for the tested runs, not a statement
that a high-probability confidence event occurs on every possible run.

For NPCB-Learned, the mean confidence set excludes at least one
regional candidate height in $37.84\%$, $5.17\%$, $0\%$, $0\%$,
and $30.11\%$ of rounds in configurations A--E, respectively.
These frequencies measure candidate exclusion, not its effect
on arm selection.

\subsection{Experimental conclusions}

The two parts support complementary conclusions.

Part~I demonstrates that the choice of risk objective can qualitatively
change the preferred route.
Under the common practical exploration rule, \texttt{KCB} ($2\sigma$) reaches the
CPT-optimal final route in every trial at the reference parameters across
the downside-trap and upside-lottery instances.
The sensitivity analysis shows that CVaR and MV behavior depends on their
tuning parameters; it does not support a parameter-independent claim
that MV fails every rare-loss configuration.
The additional known-CDF experiments evaluate \texttt{KCB} and \texttt{CPT-BPE} using their
analyzed confidence constructions, include CPT-aware comparisons in
two-dimensional kernel and linear instances, and report regret across
multiple horizons.
These scaling diagnostics do not establish theoretical regret exponents.

Part~II isolates the statistical cost of learning the noise CDF within
the finite realizable model.
NPCB-Learned (Algorithm~\ref{alg:npcb-learned}) reaches the CPT-optimal
final route in every recorded trial, while paying a substantial,
instance-dependent cost relative to the known-law specialization.
It also achieves lower mean CPT regret than plug-in MLE in every
configuration, with final-route success of $100\%$ versus $0$--$20\%$.
The large Config~E gap illustrates the finite-sample cost of learning
rare-upside laws when the mean favors another route.
The confidence and feasibility diagnostics support the finite-class
implementation; these experiments do not verify the continuous-class
learned-CDF regret rates or their inverse-density dependence.

\paragraph{Uncertainty visualization.}
In the practical Part~I comparison, learning-curve shading shows mean
cumulative CPT regret $\pm$ one standard error across $60$ trials.
The additional Part~I scaling experiments use $30$ paired trials,
with error bars of one standard error.
In Part~II, learning-curve shading shows mean cumulative CPT regret
$\pm$ one standard deviation across $30$ paired trials.
Numerical regret tables report means $\pm$ one standard error.